\pdfoutput=1

\documentclass[11pt]{article}

\usepackage[]{acl}

\usepackage{times}
\usepackage{tikz}  
\usepackage{latexsym}
\usepackage{booktabs}
\usepackage{graphicx}
\usepackage{xspace}
\usepackage{CJKutf8}
\usepackage{multirow}
\usepackage{multicol}
\usepackage{amsbsy}
\usepackage{amssymb}
\usepackage{amsmath}
\usepackage{pgfplots}
\usepackage{natbib}
\usepackage{verbatim}
\usepackage{pifont}
\usepackage[T1]{fontenc}

\usepackage[utf8]{inputenc}

\usepackage{microtype}
\usepackage[Symbol]{upgreek}
\usepackage{enumerate}
\newenvironment{itemize*}%
 {\leftmargini=10pt\begin{itemize}%
  \setlength{\itemsep}{0pt}%
  \setlength{\parskip}{0pt}%
  }%
 {\end{itemize}}
\newenvironment{enumerate*}%
 {\begin{enumerate}%
  \setlength{\itemsep}{0pt}%
  \setlength{\parskip}{0pt}}%
 {\end{enumerate}}

\usepackage{listings}
\usepackage{enumitem}

\usepackage{pgfplots}
\usepgfplotslibrary{groupplots}
\usepackage{tikz}
\usepackage{amssymb}

\definecolor{gray}{HTML}{808080}
\definecolor{golden}{HTML}{c68864}
\definecolor{frenchred}{HTML}{C73933}
\definecolor{frenchblue}{HTML}{3171B5}
\definecolor{blue}{HTML}{628BC4}
\definecolor{red}{HTML}{CA403D}
\definecolor{orange}{HTML}{ED8D6B}
\definecolor{purple}{HTML}{8376A3}
\definecolor{lightred}{HTML}{F1A2AB}
\definecolor{lightorange}{HTML}{FFCD88}
\definecolor{lightpurple}{HTML}{D5CCEC}
\definecolor{lightblue}{HTML}{31BBF8}

\definecolor{red1}{HTML}{F1A2AB}
\definecolor{red2}{HTML}{FBE4E7}
\definecolor{green1}{HTML}{C1E77E}
\definecolor{green2}{HTML}{E8F7CF}
\definecolor{blue1}{HTML}{81BBF8}
\definecolor{blue2}{HTML}{D9EAFC}
\definecolor{orange1}{HTML}{F8B881}
\definecolor{orange2}{HTML}{FDE6D3}
\definecolor{greenblue1}{HTML}{81DFE4}
\definecolor{greenblue2}{HTML}{CEF5F7}
\definecolor{purple1}{HTML}{96A7FD}
\definecolor{purple2}{HTML}{D9DFFC}
\definecolor{yellow1}{HTML}{F5D480}
\definecolor{yellow2}{HTML}{F9EFCD}
\definecolor{pink1}{HTML}{F297CC}
\definecolor{pink2}{HTML}{FBDFEF}

\usepackage{xcolor,colortbl}

\definecolor{mygray}{gray}{.8}
\newcommand{\GG}{\cellcolor{mygray}}

\newcommand{\toolname}{\textsc{UniHD}\xspace}

\newcommand{\dataname}{{MHaluBench}\xspace}

\usepackage{xcolor} 

\definecolor{myorange}{rgb}{1,0.5,0} 

\definecolor{myblue}{RGB}{63,73,251}

\definecolor{mygreen}{RGB}{134, 174, 165}

\definecolor{mypurple}{RGB}{128,0,128}

\newcommand{\colordmark}{{\ding{52}}}

\usepackage{tikz}
\usepackage{subcaption} 
\usepackage{array}

\usepackage{tcolorbox}

\usepackage{graphicx,calc}
\usepackage{wrapfig}
\newlength\myheight
\newlength\mydepth
\settototalheight\myheight{Xygp}
\title{Unified Hallucination Detection for Multimodal Large Language Models}

\author{
Xiang Chen$^{\clubsuit\heartsuit}$\footnotemark[1], Chenxi Wang$^{\spadesuit\heartsuit}$\thanks{~~Equal contribution.}, Yida Xue$^{\spadesuit\heartsuit}$, Ningyu Zhang$^{\spadesuit\heartsuit}$\footnotemark[2], {\bf Xiaoyan Yang}$^{\diamondsuit}$
 \\
 {\bf Qiang Li}$^{\diamondsuit}$, {\bf Yue Shen}$^{\diamondsuit}$, {\bf Lei Liang}$^{\diamondsuit}$, {\bf Jinjie Gu}$^{\diamondsuit}$, \textbf{Huajun Chen}$^{\clubsuit\heartsuit}$\thanks{~~Corresponding author.}\\
 $^\clubsuit$College of Computer Science and Technology, Zhejiang University  \\
  $^\spadesuit$School of Software Technology, Zhejiang University  \\
 $^\heartsuit$Zhejiang University-Ant Group Joint Laboratory of Knowledge Graph  $^\diamondsuit$Ant Group \\
  \texttt{\{xiang\_chen,zhangningyu\}@zju.edu.cn}\\
  \raisebox{-\mydepth}{\includegraphics[height=1.6\myheight]{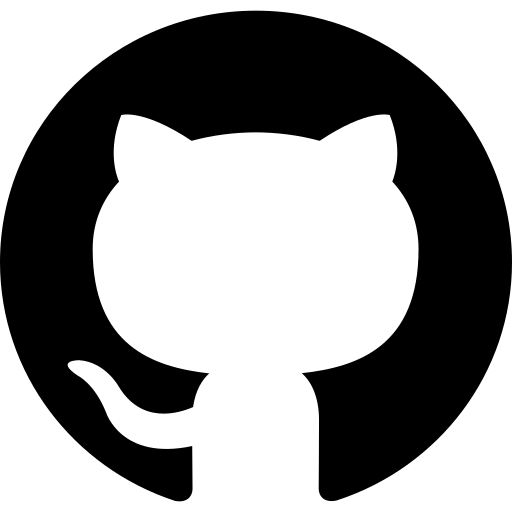}}
\textbf{\url{https://www.zjukg.org/project/EasyDetect/}}
}

\begin{document}
\maketitle
\begin{abstract}

Despite significant strides in multimodal tasks, Multimodal Large Language Models (MLLMs) are plagued by the critical issue of hallucination. The reliable detection of such hallucinations in MLLMs has, therefore, become a vital aspect of model evaluation and the safeguarding of practical application deployment. Prior research in this domain has been constrained by a narrow focus on singular tasks, an inadequate range of hallucination categories addressed, and a lack of detailed granularity. In response to these challenges, our work expands the investigative horizons of hallucination detection. We present a novel meta-evaluation benchmark, \textbf{{\dataname}}, meticulously crafted to facilitate the evaluation of advancements in hallucination detection methods. Additionally, we unveil a novel unified multimodal hallucination detection framework, \textbf{\toolname}, which leverages a suite of auxiliary tools to validate the occurrence of hallucinations robustly. We demonstrate the effectiveness of \textbf{\toolname} through meticulous evaluation and comprehensive analysis. We also provide strategic insights on the application of specific tools for addressing various categories of hallucinations\footnote{The code can be accessed via \url{https://github.com/zjunlp/EasyDetect}, and the demonstration is available at \url{http://easydetect.openkg.cn}.}.




\end{abstract}

\section{Introduction}

\begin{figure}[htbp!] 
\centering 
\includegraphics[width=0.5\textwidth]{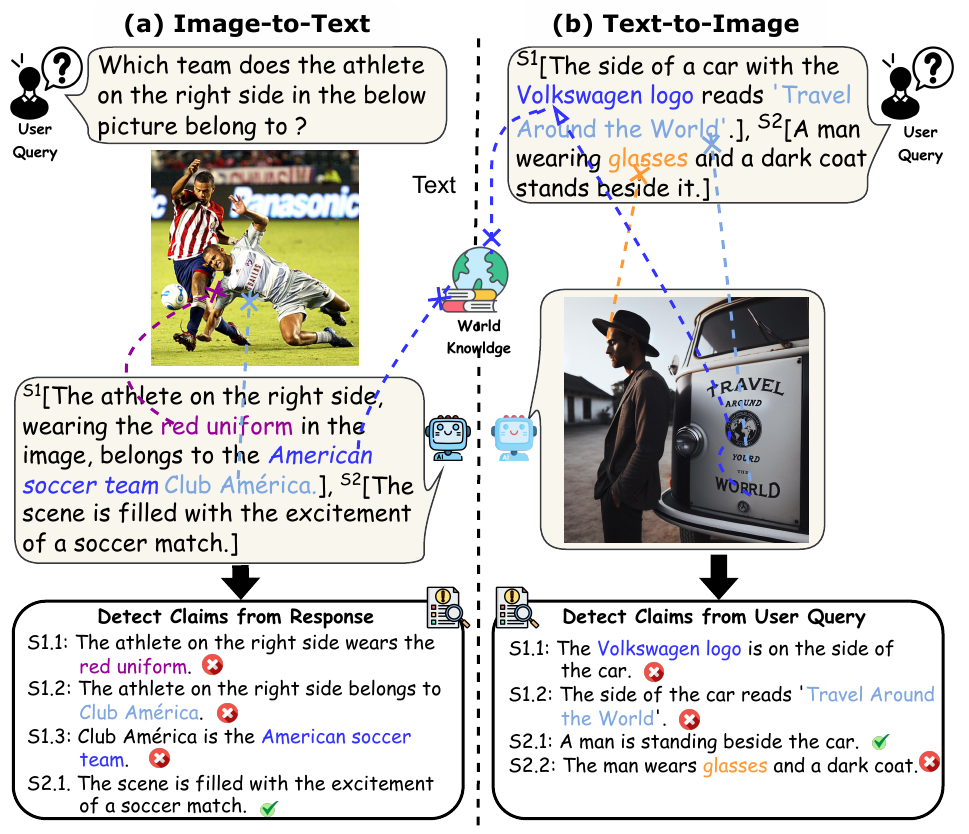} %
\caption{\small Unified multimodal hallucination detection aims to identify and detect modality-conflicting hallucinations at various levels such as \textcolor{myorange}{object}, \textcolor{mypurple}{attribute}, and \textcolor{mygreen}{scene-text}, as well as \textcolor{myblue}{fact-conflicting} hallucinations in both image-to-text and text-to-image generation. Our benchmark emphasizes fine-grained detection, with ``S1'' representing the segment and ``S1.1'' and ``S1.2'' denoting its corresponding claims.}
\label{fig:intro}
\end{figure}

The recent emergence of MLLMs~\cite{diffusion_model,gpt4,durante2024agent} that more closely mirror human cognition and learning has unleashed unprecedented possibilities for the future of artificial general intelligence (AGI).
Despite MLLMs' impressive abilities, they are susceptible to generating seemingly credible content that contradicts input data or established world knowledge, a phenomenon termed ``hallucination''\citep{liu2024survey,DBLP:journals/xihu/hallu_survey,DBLP:journals/zhongkeu/OPERA,tonmoy2024comprehensive,induce2024}.
These hallucinations hinder the practical deployment of MLLMs and contribute to the dissemination of misinformation.  Consequently, detectors that could detect multimodal hallucinations~\cite{detect23_survey} within responses from MLLMs are urgently needed to alert users to potential risks and drive the development of more reliable MLLMs.

Although several works have been conducted to detect hallucinations from MLLMs\cite{LURE,zhai2023halle,object_hallucination,HaELM} or alleviate hallucinations\cite{EFUF2024,junfei_logical2024}, these efforts operate in isolation 
and have certain limitations when compared with the aspects illustrated in Figure~\ref{fig:intro}:
\emph{(1) Task Singularity:} Current research has primarily concentrated on specific tasks, such as image captioning while neglecting that text-to-image generation, an important component of AGI, also suffers from hallucinations induced by MLLMs.
\emph{(2) Limited Hallucination Categories:} Prior studies have focused on identifying hallucinations at the object level, yet they fail to consider the prevalence of scene-text or factual inconsistencies that also frequently occur in MLLMs.
\emph{(3) Incomplete Granularity:} It would be more valuable to assess hallucinations at a fine-grained level, examining individual claims within a response, rather than evaluating the entire response holistically.
Considering these constraints hinder rapid progress in practical hallucination detection, it raises the question: \textit{Can we develop a unified perspective for detecting hallucinations from MLLMs?}

To further investigate this problem, we have broadened the concept of multimodal hallucination within MLLMs to a holistic framework, integrating both image-to-text generation such as Image Captioning (\textbf{IC}) and Visual Question Answering (\textbf{VQA}), as well as text-to-image-synthesis (\textbf{T2I}) – to align with MLLMs’ capabilities of performing varied multimodal tasks. 
We are committed to exploring a broad spectrum of hallucinatory categories and the intricate nuances of claim-level hallucination through a lens that integrates both modality-conflicting and fact-conflicting hallucinations. 
Based on the outlined perspectives,
We have developed the  \textbf{M}ultiModal \textbf{Ha}l\textbf{lu}cination Detection \textbf{Bench}mark (\textbf{\dataname})  to assess the progress of unified multimodal hallucination detectors for MLLMs and embodied the data framework depicted in Figure~\ref{fig:intro}.

At its core, leveraging MLLMs' inherent self-detection mechanisms to pinpoint diverse hallucinations encounters significant hurdles. 
We further develop a tool-augmented framework for unified hallucination detection, named \textbf{\toolname}, which integrates evidence from multiple auxiliary tools through the following procedure:
(1) \emph{Essential Claim Extraction} involves extracting the core claims within the generated response for image-to-text generation or user queries in text-to-image generation;
(2) \emph{Autonomous Tool Selection via Query Formulation} prompts MLLMs (GPT-4/Gemini) to autonomously generate pertinent questions for each claim.
These questions are crafted to determine the specific type of tool required for each claim and to establish the input for the tool's operation;
(3) \emph{Parallel Tool Execution} deploys a suite of specialized tools to operate concurrently, providing evidence from their outputs to reliably validate potential hallucinations;
(4) \emph{Hallucination Verification with Rationales} aggregates the collected evidence to instruct the underlying MLLM to judge whether the claim  hallucinatory with rationals for explanation.


We have conducted a thorough evaluation of the \textbf{\toolname} framework, utilizing the underlying MLLM against the {\dataname} benchmark. Our findings underscore the effectiveness of our approach and confirm that multimodal hallucination detection remains a formidable challenge. In a nutshell,
We conclude our contributions as:
\begin{itemize*}
    \item 
We propose a more unified problem setting for hallucination detection in MLLMs, encompassing a broad spectrum of multimodal tasks and hallucination categories, thus enriching the unified understanding of hallucination in MLLMs.

   \item 
We unveil \textbf{{\dataname}}, a meta-evaluation benchmark that encompasses various hallucination categories and multimodal tasks. This benchmark is equipped with fine-grained analytical features, gauging the progress of hallucination detectors.
 
    \item 
We introduce \textbf{\toolname}, a task-agnostic, tool-enhanced framework for the detection of hallucinations in content produced by MLLMs. Our extensive experiments demonstrate the efficacy of this method, underscoring that {\dataname} continues to be a challenging yet vital task.
\end{itemize*}
\begin{figure}[htb!] 
\centering 
\includegraphics[width=0.5\textwidth]{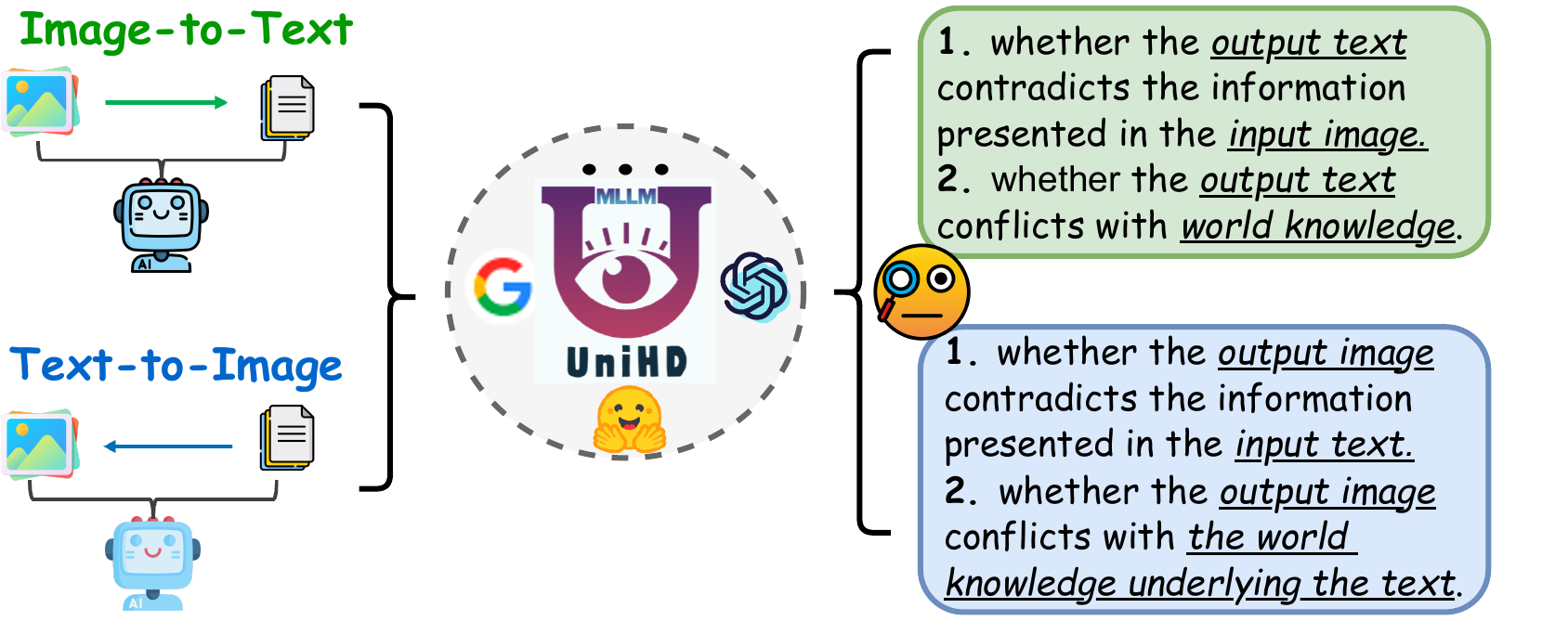} %
\caption{Unified multimodal hallucination detection.}
\label{fig:problem}
\end{figure}


\section{Preliminaries}

\begin{table*}[ht]

  \small
    \centering
   \scalebox{0.73}{
    \begin{tabular}{lccccccccc}
    \hline
    \toprule
    \multicolumn{1}{c}{\multirow{2}[4]{*}{\textbf{Datasets}}} 
    & \multicolumn{1}{c}{\textbf{Response}} 
     & \textbf{Purpose} 
    & \textbf{Granularity} 
    & \multicolumn{4}{c}{\textbf{Hallucination Types}} 
     & \textbf{Modality} 
    & \multicolumn{1}{c}{\textbf{Scenario}} \\
  \cmidrule{5-8}       
 & \textbf{Generated by} & &
& \textbf{Object} & \textbf{Attribute} & \textbf{Scene Text} & \textbf{Fact} &
& \multicolumn{1}{c}{\textbf{Task}}  \\
    \midrule
    FactCC~\cite{kryscinski-etal-2020-evaluating}& Synthetic & Check.   & Sentence       & & & & \colordmark & Text & Text2Text \\
    QAGS~\cite{wang-etal-2020-asking}  & Model & Check.  & Summary      &   & & &\colordmark & Text & Text2Text  \\
    HaluEval~\cite{wang-etal-emnlp2023/HaluEval}  & ChatGPT & Det.  & Response     & & & & \colordmark & Text  & Text2Text\\
    POPE~\cite{object_hallucination}  & - & Eval. & Response     &  \colordmark & & &  & Multi. & Image2Text\\
   
    HaELM~\cite{HaELM}   & - & Det. & Response     &  &  & & & Multi.   & Image2Text \\
    
    AMBER~\cite{wang-etl-AMBER}  & - & Eval. & Response     & \colordmark & \colordmark & & & Multi.   & Image2Text \\

    \midrule
    \multirow{1}{*}{\dataname (Ours)}   & MMLMs & Det. & Res.,Seg.,Claim    & \colordmark  & \colordmark   & \colordmark   & \colordmark  & Multi.   & Image2Text/Text2Image \\

    \bottomrule
    \hline
    \end{tabular}%
    }
      \caption{
      \small
      A comparison of benchmarks w.r.t existing fact-checking or hallucination evaluation. 
      ``Check.'' indicates verifying factual consistency, ``Eval.'' denotes evaluating hallucinations generated by different LLMs, and its response is based on different LLMs under test, while ``Det.'' embodies the evaluation of a detector's capability in identifying hallucinations.
 }
  \label{tab:comparisons}%
\end{table*}%

 We explore a unified perspective on hallucination in MLLMs (illustrated in Figure~\ref{fig:problem}) with the aspiration of developing a unified detection framework.

\paragraph{Unified View of Multimodal Hallucination Taxonomy.}
A prerequisite for unified detection is the coherent categorization of the principal categories of hallucinations within MLLMs. Our paper superficially examines the following Hallucination Taxonomy from a unified perspective:

\begin{itemize}[leftmargin=*,topsep=2pt]
    \setlength{\itemsep}{-4pt}
 \item 
\textbf{Modality-Conflicting Hallucination.}  MLLMs sometimes generate outputs that conflict with inputs from other modalities, leading to issues such as incorrect objects, attributes, or scene text. An example in Figure~\ref{fig:intro} (a) includes an MLLM inaccurately describing an athlete's uniform color, showcasing an attribute-level conflict due to MLLMs' limited ability to achieve fine-grained text-image alignment.
\item 
\textbf{Fact-Conflicting Hallucination.} Outputs from MLLMs may contradict established factual knowledge. Image-to-text models can generate narratives that stray from the actual content by incorporating irrelevant facts, while text-to-image models may produce visuals that fail to reflect the factual knowledge contained in text prompts. These discrepancies underline the struggle of MLLMs to maintain factual consistency, representing a significant challenge in the domain.
\end{itemize}

\paragraph{Unified Detection Problem Formulation.}

Unified detection of multimodal hallucination necessitates the check of each image-text pair $a=\{v, x\}$, wherein $v$ denotes either the visual input provided to an MLLM, or the visual output synthetic by it. Correspondingly, $x$ signifies the MLLM's generated textual response based on the $v$ or the textual user query for synthesizing $v$. Within this task, each $x$ may contain multiple claims, denoted as $\{c_i\}_{i = 1 \cdots n}$. The objective for hallucination detectors is to assess each claim from $a$ to determine whether it is ``hallucinatory'' or ``non-hallucinatory'', providing a rationale for their judgments based on the provided definition of hallucination. 
Text hallucination detection from LLMs denotes a sub-case in this setting, where $v$ is null.

\section{Construction of {\dataname}  }

\begin{figure}[htbp!] 
\centering 
\includegraphics[width=0.25\textwidth]{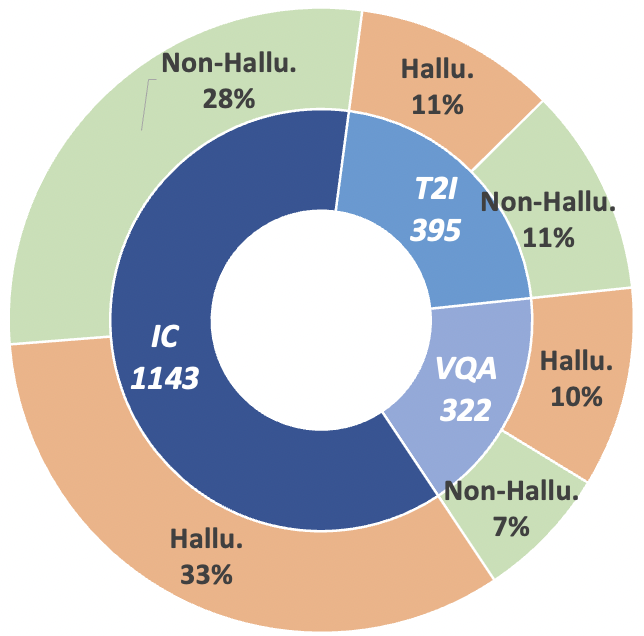} %
\caption{Claim-Level data statistics of {\dataname}. The claims are fine-grained atoms extracted from the complete ``Query-Response'' pairs. }
\label{fig:distribution1}
\end{figure}

\begin{figure}[htb!] 
\centering 
\includegraphics[width=0.35\textwidth]{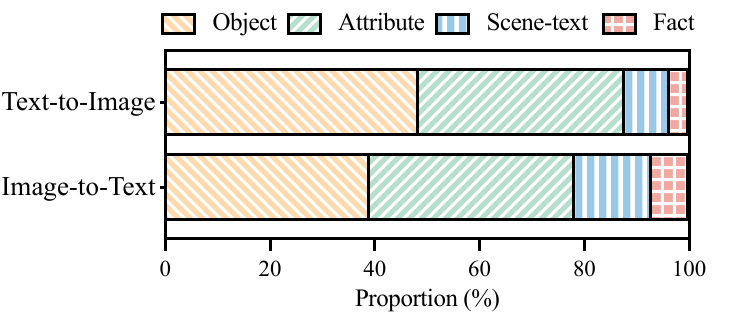} %
\caption{Distribution of hallucination categories within hallucination-labeled claims of {\dataname}.}
\label{fig:distribution2}
\end{figure}

To facilitate research in this area, we introduce the meta-evaluation benchmark {\dataname}, which encompasses the content from image-to-text and text-to-image generation, aiming to rigorously assess the advancements in multimodal hallucination detectors. 
Our benchmark has been meticulously curated to include a balanced distribution of instances across three pivotal tasks, which encompasses 200 exemplars for the task of IC  200 for VQA, and an additional 220 dedicated to Text-to-Image Generation. The comparison of {\dataname} with other benchmarks is detailed in Table~\ref{tab:comparisons} and the statistical details are provided in Figure~\ref{fig:distribution1} and Figure~\ref{fig:distribution2}.

\subsection{Hallucinatory Example Collection}
\label{subsec:example_collect}

\paragraph{Image-to-Text Generation.} 
We focus on IC and VQA tasks, drawing samples from the MS-COCO 2014 validation set~\cite{lin2014microsoft} and the TextVQA test set~\cite{TextVQA}. We compile generative outputs from mPLUG~\cite{mPLUG-Owl}, LLaVA~\cite{LLaVA}, and MiniGPT-4~\cite{MiniGPT4} to form the core dataset for {\dataname}. These models are representative of current leading MLLMs, characterized by their diverse content generation capabilities and a notable presence of hallucinations, as depicted in Figure~\ref{fig:bench_chatbot}.

\paragraph{Text-to-Image Generation.}
We source initial captions from DrawBench~\cite{DrawBench} and T2I-CompBench~\cite{T2I-CompBench}. These captions are augmented through ChatGPT to include more specific information such as objects, attributes, and factual details, among others. The refined caption guides the DALL-E 2~\cite{DALLE2} and DALL-E 3 model~\cite{BetkerImprovingIG} in producing visually detailed images.

\subsection{Segment and Claim Extraction}
\label{subsec:annotation}

Beyond evaluating overall responses, we introduce segmentation at both the segment and claim levels for a multi-granular assessment of hallucinations, enabling more precise feedback to improve model performance~\cite{lightman2023lets}. 
We leverage ChatGPT's advanced instruction-following ability to extract detailed segments and related claims. For image-to-text tasks, we split and extract the model's textual output into segments and claims; for text-to-image cases, we break down user queries into fundamental intent concepts, which are subsequently regarded as claims.





\subsection{Human Annotation and Agreement.}
\label{subsec:human_annotation}
Our annotation criteria evaluate whether image-to-text output conflicts with the input image or world knowledge and whether text-to-image visuals conflict with claims or world knowledge. Extracted claims are labeled as hallucinatory or non-hallucinatory, with a segment deemed hallucinatory if it contains any such claim; otherwise, it is labeled non-hallucinatory. An entire response is labeled hallucinatory if it includes even one hallucinatory segment.
We allocate the dataset uniformly across three annotators with graduate-level qualifications for independent categorization. Decisions in uncertain cases were initially held by individual annotators and later resolved by majority rule. Inter-annotator reliability, measured by Fleiss's Kappa (\(\kappa\)), shows significant agreement (\(\kappa = 0.822 \)) over the full annotated dataset, indicating a high level of concordance within the range \(0.80 \leq \kappa \leq 1.00\).



\section{{\toolname}: Unified Hallucination Detection Framework for MLLMs}
\label{sec:method}

We present \textbf{\toolname}  in Figure~\ref{fig:framework} and follow. The specific prompts are listed in Appendix~\ref{sec:app_prompt}


\subsection{Essential Claim Extraction}
\label{sec:method_1}
To identify fine-grained hallucinations within the response, claim extraction is a prerequisite. 
Following the procedure in \S\ref{subsec:annotation}, we employ the advanced instruction-following abilities of MLLMs for efficient claim extraction.
Specifically, GPT-4V/Gemini is adopted as the base LLM to efficiently derive verifiable claims from the outputs of image-to-text models (extracting each response into individual claims) and text-to-image models (deconstructing user queries into distinct claims)~\footnote{In subsequent experiments, our framework builds upon the pre-annotated claims available in {\dataname}, and the claim extraction is only necessary in the open-domain setting.}.



\subsection{Autonomous Tool Selection Via Query Formulation}
\label{sec:method_2}

\begin{figure*}[htb!] 
\centering 
\includegraphics[width=1.0\textwidth]{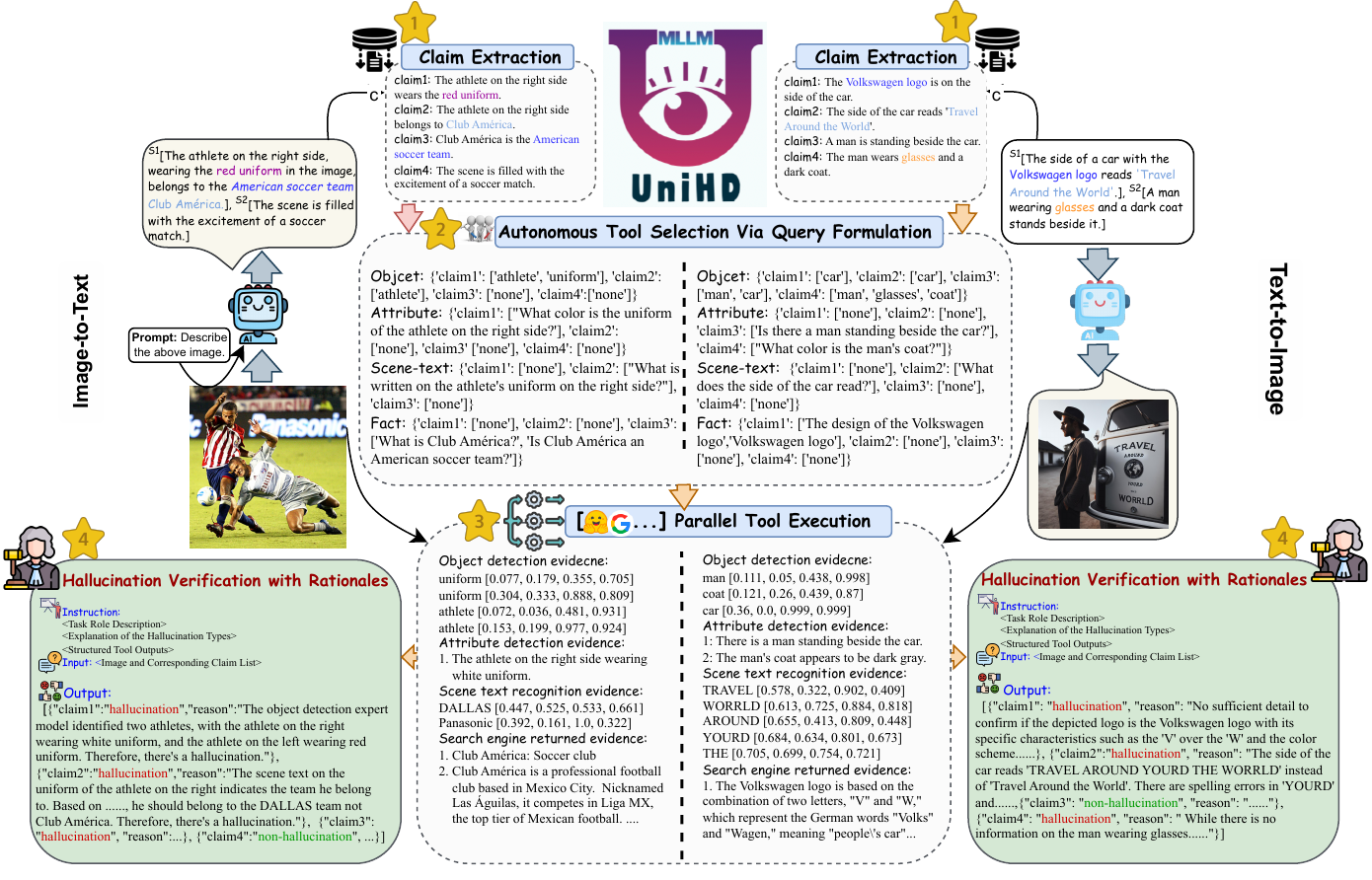} %
\caption{The specific illustration of {\toolname} for unified multimodal hallucination detection.}
\label{fig:framework}
\end{figure*}

After extracting essential claims from the input image-text pair $a=\{v, x\}$, the challenge of hallucination detection is to aptly match each claim with appropriate aspect-oriented tools. 
We approach this issue by assessing whether the underlying MLLMs can generate pertinent queries for a given set of claims $\{c_i\}_{i = 1 \cdots n}$ to provide relevant input to the specific aspect-oriented tool.
To facilitate this, we prompt underlying MLLMs like GPT-4V/Gemini to autonomously formulate meaningful queries. Demonstrated in Figure~\ref{fig:framework}, this module yields custom queries for each claim, or ``\texttt{none}'' when a tool is unnecessary. For example, the framework determines that claim1 calls for the attribute-oriented question ``\texttt{What color is the uniform of the athlete on the right side?}'' and the object-oriented inquiry ``[`\texttt{athlete}', `\texttt{uniform}']'', bypassing the need for scene-text and fact-oriented tools.

\subsection{Parallel Tool Execution}
\label{sec:method_3}
Leveraging queries autonomously generated from various perspectives, we simultaneously deploy these tools in response to the queries, gathering a comprehensive array of insights to underpin the verification of hallucinations.
The specific tools employed in our framework are detailed below, selected for their ability to effectively address a wide range of multimodal hallucination scenarios:

\begin{itemize}[leftmargin=*,topsep=2pt]
    \setlength{\itemsep}{-4pt}
 \item 
 \emph{ Object-oriented tool:} We employ the open-set object detection model Grounding DINO~\cite{dino23} for capturing visual object information, crucial for detecting object-level hallucinations. For instance, inputting ``[`\texttt{athlete}', `\texttt{uniform}']'' prompts the model to return two uniform objects and two athlete objects, along with their normalized location coordinates.
 \item 
 \emph{ Attribute-Oriented Tool:} Dealing with attributes such as positions, colors, and actions, we harness underlying MLLMs (such as GPT-4V and Gemini)  to answer the specific attribute-level questions. These responses are leveraged for hallucination verification within the same MLLMs, mirroring a self-reflect akin to \cite{shinn2023reflexion}.
  \item 
\emph{ Scene-Text-Oriented Tool:} 
Should the generated questions for scene text not be exclusively ``\texttt{none}'', we then invoke MAERec~\cite{jiang2023revisiting} as our scene-text detection tool, which is capable of identifying scene text within images along with their corresponding normalized four-dimensional coordinates.
  \item 
 \emph{ Fact-Oriented Tool:} 
 To validate conflicting factual hallucinations, we harness the Serper Google Search API to perform web searches using specific fact-based questions. By extracting and scrutinizing the top results, we obtain a range of snippets from the API's responses for analysis.
\end{itemize}

\noindent
Moreover,  {\toolname} is tool-agnostic, facilitating the seamless integration of emerging tools and detection strategies to amass tool knowledge, thereby bolstering the process of hallucination verification.
\vspace{-0.6cm}

\subsection{Hallucination Verification with Rationales }
\label{sec:method_4}
In the concluding phase of our process, we subject each claim, denoted as $c_i$, to a binary prediction to ascertain its hallucinatory status. Claims are categorized as either \textsc{Hallucinatory} or \textsc{Non-hallucinatory} based on the level of evidence support.
To accomplish this, we aggregate the collected evidence from tools with the original image and its corresponding claim list\footnote{\noindent Note that the set $a=\{v, x\}$, corresponding to the list of claims, is input into the detectors in a single batch. This operation allows the detectors to capture contextual information while also enhancing efficiency.} into a comprehensive prompt. Subsequently, we instruct our chosen MLLM (GPT-4V or Gemini) to assess each claim's hallucinatory potential. In doing so, the MLLM also generates insightful explanations to elucidate the rationale behind its judgment.

\begin{table*}[!htb]
\renewcommand\arraystretch{1.22}
  \centering
  \footnotesize
  \scalebox{0.75}{
    \begin{tabular}{c|cccccc|ccc|cccc}
    \toprule
    \multicolumn{1}{c|}{\multirow{2}[2]{*}{\textbf{Tasks}}} & \multicolumn{1}{r}{\multirow{2}[2]{*}{\textbf{LLMs}}} 
    & \multirow{2}[2]{*}{\textbf{Methods}}
    & \multirow{2}[2]{*}{\textbf{Levels}}
    & \multicolumn{3}{c}{\textbf{Hallucinatory}} 
    & \multicolumn{3}{c}{\textbf{Non-Hallucinatory}} 
    & \multicolumn{4}{c}{\textbf{ Average}}\\
\cline{5-14}           &       &       &  
& \textbf{P} & \textbf{R} & \textbf{F1} 
& \textbf{P} & \textbf{R} & \textbf{F1} 
& \textbf{Acc.} & \textbf{P} & \textbf{R} & \textbf{Mac.F1} \\
    \midrule
    \multicolumn{1}{c|}{\multirow{12}[6]{*}{Image-to-Text}} & \multicolumn{1}{c}{\multirow{6}[0]{*}{Gemini}} 
    &\multicolumn{1}{c}{\multirow{2}[0]{*}{Self-Check (0-shot)}} & Claim &  83.17  &  42.15 & 55.95 &
    55.64 & 89.48 & 68.61 & 63.34  & 69.41 & 65.82 &   62.28\\
     &  &   & Segment  &  89.30    & 47.71  &  62.19 & 43.76 &  87.68  & 58.38 & 60.38 &  66.53 & 67.69 & 60.29\\
     \cline{4-14}
    &   & \multicolumn{1}{c}{\multirow{2}[0]{*}{Self-Check (2-shot)}} 
 & Claim  & 84.24 & 66.75 & 74.48 & 67.35 & 84.60 & 75.00 & 74.74 & 75.80 & 75.68 & 74.74  \\
    &   &     & Segment  & 90.44 & 71.08 & 79.60 & 57.35 &   83.80 & 68.10 & 75.11 & 73.89 & 77.44 & 73.85 \\
    \cline{4-14}
    &       & \multicolumn{1}{c}{\multirow{2}[0]{*}{\textbf{\toolname}}}   & Claim & 84.44 & 72.44 &   77.98 & 71.08 & 83.54 & 76.80 & 77.41 & 77.76 &  77.99 & 77.39 \\
    &       &  & Segment & 88.77 & 78.76 & 83.46 & 63.17 & 78.52 & 70.02 & 78.68 & 75.97 & 78.64 & 76.74 \\

\cline{2-14}     
&  \multicolumn{1}{c}{\multirow{6}[0]{*}{GPT-4v}} 
    &\multicolumn{1}{c}{\multirow{2}[0]{*}{Self-Check (0-shot)}} & Claim  & 79.37 & 74.17 & 76.68 & 70.52 & 76.22 & 73.26 & 75.09 & 74.94 & 75.19 & 74.97\\
     &  &   & Segment & 84.78 &  80.07 & 82.35 & 61.64 & 69.01 & 65.12 & 76.56 & 73.21 & 74.54 & 73.73\\
     \cline{4-14}
    &   & \multicolumn{1}{c}{\multirow{2}[0]{*}{Self-Check (2-shot)}} 
 & Claim  &  82.00 & 79.98 & 80.98 & 76.04 & 78.35 & 77.18 & 79.25 & 79.02  & 79.16 & 79.08 \\
    &   &     & Segment  & 86.54 & 85.13 & 85.83 & 69.05 &  71.48 & 70.24 & 80.80 & 77.80 & 78.30 & 78.04 \\
     \cline{4-14}
    &       & \multicolumn{1}{c}{\multirow{2}[0]{*}{\textbf{\toolname}}}   & \GG Claim & \GG 82.54 & \GG 85.29 & \GG 83.89 & \GG 81.08 & \GG 77.74 & \GG 79.38 & \GG 81.91 & \GG 81.81 & \GG 81.52 & \GG 81.63 \\
    &       &  & \GG Segment & \GG 87.03 & \GG 91.01 & \GG 88.98  & \GG 78.52 & \GG 70.77 &  \GG 74.44 & \GG 84.60 & \GG 82.77 & \GG 80.89 & \GG 81.71 \\

\midrule
    \multicolumn{1}{c|}{\multirow{12}[6]{*}{Text-to-Image}} & \multicolumn{1}{c}{\multirow{6}[0]{*}{Gemini}} 
    &\multicolumn{1}{c}{\multirow{2}[0]{*}{Self-Check (0-shot)}} & Claim  & 73.85 &  24.62 & 36.92 
    & 55.45  & 91.50 & 69.06 & 58.48 & 64.65 & 58.06 & 52.99 \\
     &  &   & Segment  & 87.27 & 30.00 & 44.65 & 32.53 & 88.52 & 47.58 & 46.15 & 59.90 & 59.26 & 46.11 \\
    \cline{4-14}
    &   & \multicolumn{1}{c}{\multirow{2}[0]{*}{Self-Check (2-shot)}} 
 & Claim  & 85.37 & 53.85 & 66.04 & 66.91 & 91.00 & 77.12 & 72.66 & 76.14 & 72.42 & 71.58   \\
    &   &     & Segment  & 91.67 & 61.88  & 73.88 & 46.02 & 85.25 & 59.77 & 68.33 & 68.84 & 73.56 & 66.83  \\
    \cline{4-14}
    &       & \multicolumn{1}{c}{\multirow{2}[0]{*}{\textbf{\toolname}}}   & Claim & 85.71 & 61.54 & 71.64 & 70.59 &   90.00 & 79.12 & 75.95 & 78.15 &   75.77 & 75.38    \\
    &       &  & Segment & 93.28 & 69.37 & 79.57 & 51.96 & 86.89 & 65.03 & 74.21 & 72.62 & 78.13 &   72.30\\
\cline{2-14}     
&  \multicolumn{1}{r}{\multirow{6}[0]{*}{GPT-4v}} 
    &\multicolumn{1}{c}{\multirow{2}[0]{*}{Self-Check (0-shot)}} & Claim  & 88.55 & 59.49 & 71.17 & 70.08 & 92.50 & 79.74 & 76.20 & 79.31 & 75.99 &  75.45   \\
     &  &   & Segment  & 93.69   & 65.00 & 76.75 & 49.09 & 88.52 & 63.16 & 71.49 & 71.39 & 76.76 & 69.96  \\
    \cline{4-14}
    &   & \multicolumn{1}{c}{\multirow{2}[0]{*}{Self-Check (2-shot)}} 
 & Claim  & 84.39 & 74.87   & 79.35 & 77.93 & 86.50 & 81.99 & 80.76 & 81.16 & 80.69 & 80.67   \\
    &   &     & Segment  & 89.63 & 75.62 & 82.03 & 54.65 & 77.05 & 63.95 & 76.02 & 72.14 & 76.34 &   72.99\\
    \cline{4-14}
    &       & \multicolumn{1}{c}{\multirow{2}[0]{*}  {\textbf{\toolname}}}   & \GG  Claim & \GG 84.92 & \GG 86.67 & \GG 85.79  &\GG 86.73 & \GG 85.00  & \GG 85.86 &\GG 85.82 & \GG 85.83 &\GG  85.83 &\GG  85.82 \\
    &       &  & \GG Segment & \GG 91.25   & \GG 91.25 & \GG 91.25  & \GG 77.05& \GG 77.05 & \GG 77.05 & \GG 87.33 & \GG 84.15 & \GG 84.15 & \GG 84.15\\
    \bottomrule
    \end{tabular}
     }
      \caption{Experimental results of {\toolname} powered by Gemini and GPT-4V on Image-to-Text and Text-to-Image Generation. The default F1 score is Micro-F1, whereas Mac.F1 represents the Macro-F1 score.}
  \label{tab:allresults}%
\end{table*}%

\section{Experiment}
\label{sec:experiment}

\subsection{Experimental Settings}

\paragraph{Baselines.} We compare {\toolname} on {\dataname}\footnote{In this paper, we conducted experiments using the evaluation benchmark from our published V0.1 version.} with two baselines, Self-Check (2-shot)\footnote{Self-Check (2-shot) utilize two complete demonstrations based on $a=\{v, x\}$ rather than only two claims.} and Self-Check (0-shot) based on CoT~\cite{wei2023chainofthought}, which assess the capability of the underlying MLLM to identify hallucinations without external knowledge and have shown effectiveness across other various tasks~\cite{factool,xie2023ask}.
We prompt GPT-4V (\texttt{gpt-4-vision-preview}) and Gemini (Pro Vision)
to recognize fine-grained hallucinations and explain the reasoning behind this determination.

\paragraph{Evaluation Perspective.}

We compute the recall, precision, and Micro-F1 metrics individually for both hallucinatory and non-hallucinatory categories. Additionally, we assess the overall performance by measuring the average Macro-F1 scores at the claim and segment levels. We categorize a segment as non-hallucinatory only if all associated claims are classified as non-hallucinatory; it is deemed hallucinatory if any associated claims do not meet this criterion.

\subsection{Evaluation Results}
\label{sec:result}

\paragraph{\dataname poses a challenging benchmark for multimodal hallucination detection.} 

The segment-level and response-level outcomes are presented in Table~\ref{tab:allresults}. 
Even though all hallucinatory instances in \dataname are obtained from open-source MLLMs' outputs rather than being generated by GPT-4V/Gemini itself, it is noteworthy that the majority of detectors achieve an overall Macro-F1 score ranging between 70\%-80\%, exhibiting subpar performance on \dataname.  

\input{table/ablation}

\begin{figure*}[htb!] 
\centering 
\includegraphics[width=0.95\textwidth]{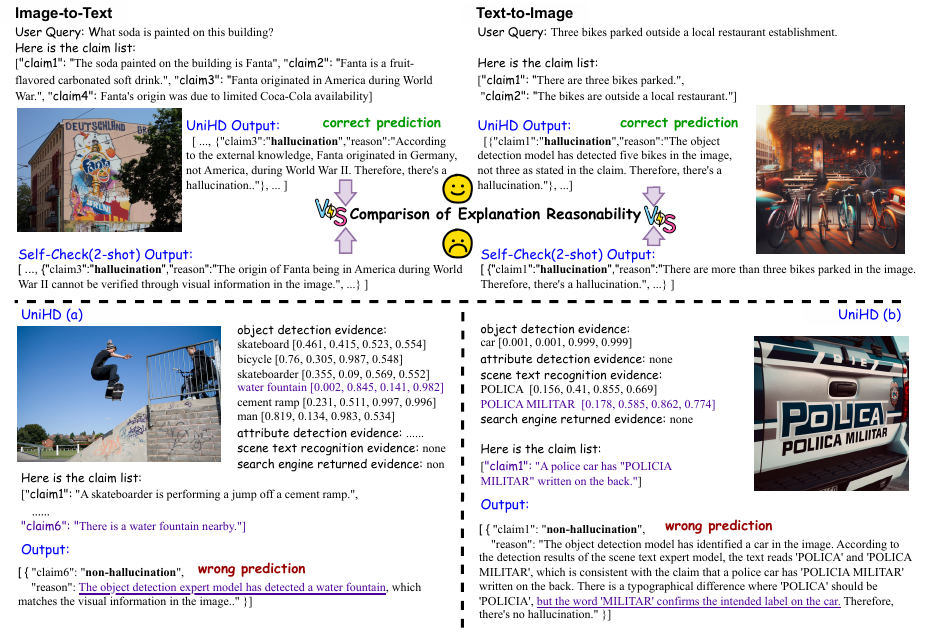} %
\caption{
\textbf{Case Study.} The upper section depicts two exemplary cases where both {\toolname} and Self-Check (2-shot) arrive at correct judgments, with a comparative demonstration of {\toolname} providing explanations of superior reasonability.  {\toolname} (a) reveals a failure case where the tool presents erroneous evidence, leading to an incorrect verification outcome. Conversely, {\toolname} (b) highlights a scenario where, despite the tool offering valid and correct evidence, GPT-4V persists in its original stance, resulting in a flawed verification.}
\label{fig:case_analysis}
\end{figure*}

\paragraph{GPT-4V surpasses Gemini as the detector base.}

GPT-4V-powered detectors consistently outperform Gemini counterparts, achieving higher Macro-F1 scores, especially in the text-to-image generation. For instance, Self-Check (0-shot) using GPT-4V achieves a claim-level Macro-F1 of $72.82$, significantly surpassing Gemini's Macro-F1 score of $52.98$. However, Gemini-powered detectors exhibit better performance in non-hallucinatory categories for image-to-text tasks, indicating a potential bias towards reduced sensitivity to hallucinations.

\paragraph{\toolname Empowered by GPT-4V: Superior Detection Across the Board.}
Table~\ref{tab:allresults} demonstrates that {\toolname}, leveraging GPT-4V, consistently outperforms other baseline detectors in image-to-text and text-to-image tasks. Despite the Self-Check (2-shot) showcasing GPT-4V and Gemini's robust in-context learning, {\toolname} markedly exceeds its performance, emphasizing the benefits of integrating external tools for more robust evidence verification and reliable hallucination detection.

\subsection{Analysis}
\label{sec:experimental_analysis}
\paragraph{Which Type of Hallucination Can Benefit the Most from Tool Enhancement?}
Figure~\ref{fig:statistic_hallucination} shows that {\toolname}  enhances the detection of \textbf{scene text} and \textbf{factual} hallucinations over Self-Check (2-shot), suggesting that GPT-4V or Gemini's inherent limitations make the evidence provided by the tool especially valuable.
However, {\toolname} exhibits minimal improvement in identifying \textbf{attribute-level} hallucinations, potentially attributed to a lack of specialized tools for direct attribute detection, with self-reflection methods based on GPT-4V/Gemini proving to be relatively weak.

\paragraph{Explanation Reasonability of \toolname.}
As shown in the upper portion of Figure~\ref{fig:case_analysis}, 
 both the fact-level hallucination ``Fanta originated in America during World War.'' and the object-level hallucination ``There are three bikes parked.'' are accurately identified by Self-Check (2-shot) and \toolname. Comparative analysis reveals that \toolname excels in synthesizing evidence to provide a more credible and compelling rationale.

\paragraph{Failure Analysis of \toolname.}
As shown in the lower part of Figure~\ref{fig:case_analysis}, we present two instances where \toolname exhibits limitations. The left case demonstrates situations where the tool either generates incorrect evidence or fails to provide useful information, leading to erroneous judgments by the MLLM. On the right, we observe cases where the MLLM maintains its initial bias despite receiving accurate evidence, resulting in incorrect decisions.
\begin{figure}[htb!] 
\centering 
\includegraphics[width=0.48\textwidth]{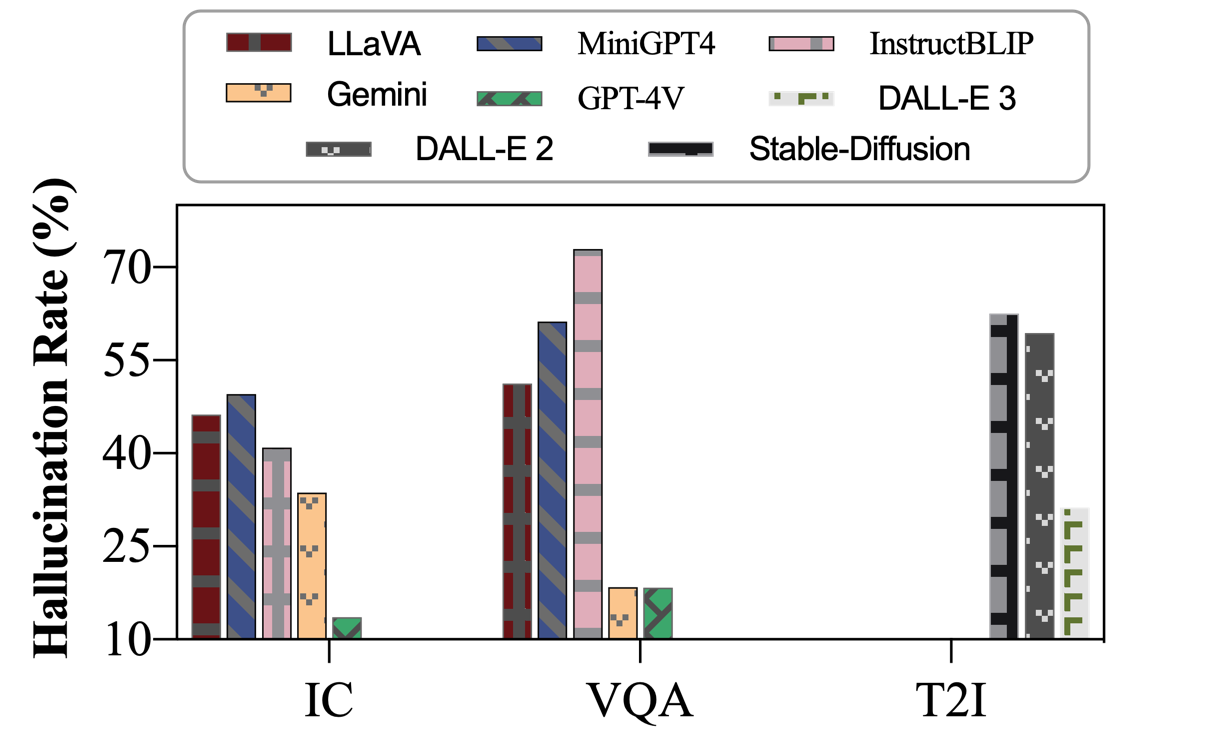} %
\caption{
Comparison of claim-level hallucination ratios across MLLMs.
We randomly select a set of 20 prompts from {\dataname} for each of the IC, VQA, and T2I.
Responses for these prompts are generated by each of the evaluated MLLMs.}
\label{fig:bench_chatbot}
\vspace{-0.5cm}
\end{figure}
These scenarios highlight areas for further research to enhance tool accuracy and to develop MLLMs dedicated to better hallucination detection.

\paragraph{Text-to-Image Hallucination vs. Image-to-Text Hallucination: Which is Easier to Detect?}
Both baselines and the GPT-4V-enhanced \toolname show significantly improved performance in identifying hallucinations in text-to-image content over image-to-text content. This can be traced back to the structured nature of manually written user queries for text-to-image tasks, which yield more uniform images.
while image-to-text confronts the complexity of natural images with background noise and content generated by MLLMs, characterized by greater diversity and fewer constraints. Consequently, it is intuitively easier to detect discrepancies between text and corresponding images in text-to-image tasks.

\paragraph{Explore {\toolname} to Evaluate Hallucination of Modern MLLMs.}
We designate {\toolname} powered by GPT-4V as the golden detector to assess the frequency of hallucinations in MLLMs, including GPT-4V, and Gemini, among others. The findings illustrated in Figure~\ref{fig:bench_chatbot} indicate that (1) GPT-4V exhibits the lowest claim-level hallucination ratio across most tested conditions, and (2) the hallucination-based ranking of these MLLMs is generally in agreement with established leaderboards and human evaluation, demonstrating the potential of {\toolname} for evaluating hallucinations.


\section{Related Work}

\subsection{Hallucinations in MLLM}

The advent of MLLMs~\cite{gpt4,LLaVA,mPLUG-Owl,MiniGPT4} has highlighted the issue of hallucination~\cite{fact_iclr2024,suvery-Siren,hit23_survey,found_hallucination,23old_survey}, a crucial concern impacting their dependability. Previous research has primarily focused on three areas: evaluating~\cite{object_hallucination,HallusionBench,FaithScore}, detecting~\cite{HaELM,detect23_survey,Woodpecker}, and mitigating hallucinations~\cite{KCA_tecent,DBLP:journals/align, DBLP:journals/zhongkeu/OPERA,semnani2023wikichat,MARINE,vcd_cvpr2024,icd2024,cgd2024}.
In a complementary effort, HaELM~\cite{HaELM} scrutinizes the challenges associated with POPE~\cite{object_hallucination} and suggests training a model based on simulated hallucination samples for detecting multimodal hallucinations.
Diverging from prior efforts, this paper addresses a broader problem scope for hallucination detection, introducing a unified multimodal hallucination detection framework, {\toolname}, along with meta-evaluation benchmarks, {\dataname}.




\subsection{Harnessing Tool Resources for LLMs}

Addressing the limitations of LLMs~\cite{huajun23,CRAG} due to their pre-training confinement, researchers have explored augmenting them with resources like knowledge bases, search engines, and external models, to expand their functionality. 
Notably, \citet{schick2023toolformer,ToolkenGPT,toolfeed23} have developed models that leverage external tools to improve performance in downstream tasks. 
More recently, \citet{shen2023hugginggpt,liang2023taskmatrixai}  has unveiled frameworks integrating LLMs with diverse AI models to tackle complex challenges. Building on this, researchers~\cite{peng2023check,factchd}  have examined the utilization of external knowledge to mitigate or evaluate hallucinations in LLMs. 
Adapting these enhancements for MLLMs introduces unique challenges, necessitating the selection of appropriate tools for effective oversight. Our research focuses on automating the selection of functionally diverse tools to enhance multimodal hallucination detection.



\section{Conclusion}

We introduce a unified problem formulation for multimodal hallucination detection that encompasses a diverse range of multimodal tasks and hallucination types. A fine-grained benchmark dataset, {\dataname}, is also proposed to promote this challenging direction. Alongside this, we present the unified hallucination detection framework, \toolname, capable of autonomously selecting external tools with capturing pertinent knowledge to support hallucination verification with rationales. Our experimental results indicate that \toolname achieves better performance across both image-to-text and text-to-image generation tasks, confirming its universality and efficacy.

\section*{Limitations}
This paper focuses on constructing a unified hallucination detection framework for MLLMs, dubbed {\toolname}.
Despite the best efforts, our paper still have some limitations.

\paragraph{The Scope of Multimodal Tasks.}
This paper primarily addresses the detection of multimodal hallucinations from a unified perspective, with a focus on image-to-text tasks (such as Image Captioning and VQA) and text-to-image generation tasks. Nonetheless, it is important to recognize that our framework does not yet encompass other multimodal tasks, such as video captioning, which are also susceptible to hallucinations. Moving forward, we aim to explore the possibilities of incorporating these additional domains into our {\toolname}.

\paragraph{Limitations of Closed-Source MLLM Pricing and Inference Speed.}
Our {\toolname} is primarily built upon powerful closed-source models as the foundation. However, closed-source models~\cite{LLaVA,MiniGPT4,mPLUG-Owl,Qwen} often come with a cost, which introduces operational expenses. Additionally, our {\toolname} relies on several external tools to provide evidence for enhanced illusion verification, resulting in additional inference time. In the future, we will further explore training open-source dedicated illusion detection models with the tool to further improve effectiveness and reduce costs.

\paragraph{The Scope of Hallucination Categories.}
In our commitment to developing a comprehensive hallucination detection framework, referred to as {\toolname}, for MLLMs, we have made efforts to incorporate various prevalent hallucination categories within {\dataname} and {\toolname}, including object, attribute, scene-text, and factual aspects, among others. However, it is important to acknowledge that there are additional categories of hallucinations that have not been covered in our framework, as discussed in the existing literature~\cite{suvery-Siren, DBLP:journals/xihu/hallu_survey,mishra2024finegrained,hit23_survey,found_hallucination}
. Moving forward, our research will expand its scope to adopt a unified approach towards a wider range of hallucination categories, to strengthen the robustness of our detection mechanisms.

\paragraph{Preliminary Attempts at Tool Utilization.}
In our early endeavors, we have configured a dedicated tool for detecting a specific type of hallucination, exemplified by the assignment of the Grounded DINO model as the object detection tool of choice. However, it should be acknowledged that the current selection of tools may not represent the optimum choice. It remains imperative to rigorously explore which SOTA object detection models are best suited for the task of multimodal hallucination detection. This necessitates an extensive evaluation of available models to pinpoint the most effective tool that aligns with the nuances and complexities of our multimodal detection objectives.

\section*{Acknowledgement}
We are grateful for the API services provided by OpenAI and Google, which enabled us to process data and conduct some of our experiments.
Part implementation of this work are assisted and inspired by the related hallucination toolkits including FactTool \cite{factool}, Woodpecker \cite{Woodpecker}, and others. 
We follow the same license for open-sourcing and thank them for their contributions to the community.
This work also benefits from the public project of mPLUG-Owl\footnote{\url{https://github.com/X-PLUG/mPLUG-Owl}}, MiniGPT-4\footnote{\url{https://github.com/Vision-CAIR/MiniGPT-4}}, LLaVA\footnote{\url{https://github.com/haotian-liu/LLaVA}}, GroundingDINO\footnote{\url{https://github.com/IDEA-Research/GroundingDINO}}, and MAERec\footnote{\url{https://github.com/Mountchicken/Union14M}}. 
This work was supported by the National Natural Science Foundation of China (No. 62206246), the Fundamental Research Funds for the Central Universities (226-2023-00138), Zhejiang Provincial Natural Science Foundation of China (No. LGG22F030011), Yongjiang Talent Introduction Programme (2021A-156-G), CCF-Tencent Rhino-Bird Open Research Fund, and Information Technology Center and State Key Lab of CAD\&CG, Zhejiang University.



\bibliography{custom}

\begin{thebibliography}{61}
\expandafter\ifx\csname natexlab\endcsname\relax\def\natexlab#1{#1}\fi

\bibitem[{Bai et~al.(2023)Bai, Bai, Yang, Wang, Tan, Wang, Lin, Zhou, and Zhou}]{Qwen}
Jinze Bai, Shuai Bai, Shusheng Yang, Shijie Wang, Sinan Tan, Peng Wang, Junyang Lin, Chang Zhou, and Jingren Zhou. 2023.
\newblock \href {https://doi.org/10.48550/ARXIV.2308.12966} {Qwen-vl: {A} frontier large vision-language model with versatile abilities}.
\newblock \emph{CoRR}, abs/2308.12966.

\bibitem[{Betker et~al.(2023)Betker, Goh, Jing, TimBrooks, Wang, Li, LongOuyang, JuntangZhuang, JoyceLee, YufeiGuo, WesamManassra, PrafullaDhariwal, CaseyChu, YunxinJiao, and Ramesh}]{BetkerImprovingIG}
James Betker, Gabriel Goh, Li~Jing, TimBrooks, Jianfeng Wang, Linjie Li, LongOuyang, JuntangZhuang, JoyceLee, YufeiGuo, WesamManassra, PrafullaDhariwal, CaseyChu, YunxinJiao, and Aditya Ramesh. 2023.
\newblock \href {https://cdn.openai.com/papers/dall-e-3.pdf} {Improving image generation with better captions}.

\bibitem[{Chen(2023)}]{huajun23}
Huajun Chen. 2023.
\newblock \href {https://doi.org/10.48550/ARXIV.2312.02706} {Large knowledge model: Perspectives and challenges}.
\newblock \emph{CoRR}, abs/2312.02706.

\bibitem[{Chen et~al.(2023)Chen, Song, Gui, Wang, Zhang, Yong, Huang, Lv, Zhang, and Chen}]{factchd}
Xiang Chen, Duanzheng Song, Honghao Gui, Chengxi Wang, Ningyu Zhang, Jiang Yong, Fei Huang, Chengfei Lv, Dan Zhang, and Huajun Chen. 2023.
\newblock \href {https://doi.org/10.48550/ARXIV.2310.12086} {Factchd: Benchmarking fact-conflicting hallucination detection}.
\newblock \emph{CoRR}, abs/2310.12086.

\bibitem[{Chern et~al.(2023)Chern, Chern, Chen, Yuan, Feng, Zhou, He, Neubig, and Liu}]{factool}
I{-}Chun Chern, Steffi Chern, Shiqi Chen, Weizhe Yuan, Kehua Feng, Chunting Zhou, Junxian He, Graham Neubig, and Pengfei Liu. 2023.
\newblock \href {https://doi.org/10.48550/ARXIV.2307.13528} {Factool: Factuality detection in generative {AI} - {A} tool augmented framework for multi-task and multi-domain scenarios}.
\newblock \emph{CoRR}, abs/2307.13528.

\bibitem[{Deng et~al.(2024)Deng, Chen, and Hooi}]{cgd2024}
Ailin Deng, Zhirui Chen, and Bryan Hooi. 2024.
\newblock \href {https://doi.org/10.48550/ARXIV.2402.15300} {Seeing is believing: Mitigating hallucination in large vision-language models via clip-guided decoding}.
\newblock \emph{CoRR}, abs/2402.15300.

\bibitem[{Durante et~al.(2024)Durante, Huang, Wake, Gong, Park, Sarkar, Taori, Noda, Terzopoulos, Choi, Ikeuchi, Vo, Fei-Fei, and Gao}]{durante2024agent}
Zane Durante, Qiuyuan Huang, Naoki Wake, Ran Gong, Jae~Sung Park, Bidipta Sarkar, Rohan Taori, Yusuke Noda, Demetri Terzopoulos, Yejin Choi, Katsushi Ikeuchi, Hoi Vo, Li~Fei-Fei, and Jianfeng Gao. 2024.
\newblock \href {http://arxiv.org/abs/2401.03568} {Agent ai: Surveying the horizons of multimodal interaction}.

\bibitem[{Hao et~al.(2023)Hao, Liu, Wang, and Hu}]{ToolkenGPT}
Shibo Hao, Tianyang Liu, Zhen Wang, and Zhiting Hu. 2023.
\newblock Toolkengpt: Augmenting frozen language models with massive tools via tool embeddings.
\newblock \emph{NeurIPS 2023}.

\bibitem[{Ho et~al.(2020)Ho, Jain, and Abbeel}]{diffusion_model}
Jonathan Ho, Ajay Jain, and Pieter Abbeel. 2020.
\newblock \href {https://proceedings.neurips.cc/paper/2020/hash/4c5bcfec8584af0d967f1ab10179ca4b-Abstract.html} {Denoising diffusion probabilistic models}.
\newblock In \emph{Advances in Neural Information Processing Systems 33: Annual Conference on Neural Information Processing Systems 2020, NeurIPS 2020, December 6-12, 2020, virtual}.

\bibitem[{Hu et~al.(2024)Hu, Chen, Li, Guo, Wen, Yu, and Guo}]{fact_iclr2024}
Xuming Hu, Junzhe Chen, Xiaochuan Li, Yufei Guo, Lijie Wen, Philip~S. Yu, and Zhijiang Guo. 2024.
\newblock Do large language models know about facts?
\newblock \emph{ICLR 2024}.

\bibitem[{Huang et~al.(2023{\natexlab{a}})Huang, Sun, Xie, Li, and Liu}]{T2I-CompBench}
Kaiyi Huang, Kaiyue Sun, Enze Xie, Zhenguo Li, and Xihui Liu. 2023{\natexlab{a}}.
\newblock \href {https://doi.org/10.48550/ARXIV.2307.06350} {T2i-compbench: {A} comprehensive benchmark for open-world compositional text-to-image generation}.
\newblock \emph{CoRR}, abs/2307.06350.

\bibitem[{Huang et~al.(2023{\natexlab{b}})Huang, Yu, Ma, Zhong, Feng, Wang, Chen, Peng, Feng, Qin, and Liu}]{hit23_survey}
Lei Huang, Weijiang Yu, Weitao Ma, Weihong Zhong, Zhangyin Feng, Haotian Wang, Qianglong Chen, Weihua Peng, Xiaocheng Feng, Bing Qin, and Ting Liu. 2023{\natexlab{b}}.
\newblock \href {https://doi.org/10.48550/ARXIV.2311.05232} {A survey on hallucination in large language models: Principles, taxonomy, challenges, and open questions}.
\newblock \emph{CoRR}, abs/2311.05232.

\bibitem[{Huang et~al.(2023{\natexlab{c}})Huang, Dong, Zhang, Wang, He, Wang, Lin, Zhang, and Yu}]{DBLP:journals/zhongkeu/OPERA}
Qidong Huang, Xiaoyi Dong, Pan Zhang, Bin Wang, Conghui He, Jiaqi Wang, Dahua Lin, Weiming Zhang, and Nenghai Yu. 2023{\natexlab{c}}.
\newblock \href {https://doi.org/10.48550/ARXIV.2311.17911} {{OPERA:} alleviating hallucination in multi-modal large language models via over-trust penalty and retrospection-allocation}.
\newblock \emph{CoRR}, abs/2311.17911.

\bibitem[{Ji et~al.(2023)Ji, Lee, Frieske, Yu, Su, Xu, Ishii, Bang, Madotto, and Fung}]{23old_survey}
Ziwei Ji, Nayeon Lee, Rita Frieske, Tiezheng Yu, Dan Su, Yan Xu, Etsuko Ishii, Ye~Jin Bang, Andrea Madotto, and Pascale Fung. 2023.
\newblock \href {https://doi.org/10.1145/3571730} {Survey of hallucination in natural language generation}.
\newblock \emph{ACM Comput. Surv.}, 55(12).

\bibitem[{Jiang et~al.(2023)Jiang, Wang, Peng, Liu, and Jin}]{jiang2023revisiting}
Qing Jiang, Jiapeng Wang, Dezhi Peng, Chongyu Liu, and Lianwen Jin. 2023.
\newblock Revisiting scene text recognition: A data perspective.
\newblock In \emph{Proceedings of the IEEE/CVF international conference on computer vision}.

\bibitem[{Jing et~al.(2023)Jing, Li, Chen, Jia, and Du}]{FaithScore}
Liqiang Jing, Ruosen Li, Yunmo Chen, Mengzhao Jia, and Xinya Du. 2023.
\newblock \href {https://doi.org/10.48550/ARXIV.2311.01477} {{FAITHSCORE:} evaluating hallucinations in large vision-language models}.
\newblock \emph{CoRR}, abs/2311.01477.

\bibitem[{Kang et~al.(2024)Kang, G{\"{u}}rel, Yu, Song, and Li}]{CRAG}
Mintong Kang, Nezihe~Merve G{\"{u}}rel, Ning Yu, Dawn Song, and Bo~Li. 2024.
\newblock \href {https://doi.org/10.48550/ARXIV.2402.03181} {{C-RAG:} certified generation risks for retrieval-augmented language models}.
\newblock \emph{CoRR}, abs/2402.03181.

\bibitem[{Kryscinski et~al.(2020)Kryscinski, McCann, Xiong, and Socher}]{kryscinski-etal-2020-evaluating}
Wojciech Kryscinski, Bryan McCann, Caiming Xiong, and Richard Socher. 2020.
\newblock \href {https://doi.org/10.18653/v1/2020.emnlp-main.750} {Evaluating the factual consistency of abstractive text summarization}.
\newblock In \emph{Proceedings of the 2020 Conference on Empirical Methods in Natural Language Processing (EMNLP)}, pages 9332--9346, Online. Association for Computational Linguistics.

\bibitem[{Leng et~al.(2023)Leng, Zhang, Chen, Li, Lu, Miao, and Bing}]{vcd_cvpr2024}
Sicong Leng, Hang Zhang, Guanzheng Chen, Xin Li, Shijian Lu, Chunyan Miao, and Lidong Bing. 2023.
\newblock \href {https://doi.org/10.48550/ARXIV.2311.16922} {Mitigating object hallucinations in large vision-language models through visual contrastive decoding}.
\newblock \emph{CoRR}, abs/2311.16922.

\bibitem[{Li et~al.(2023{\natexlab{a}})Li, Cheng, Zhao, Nie, and Wen}]{wang-etal-emnlp2023/HaluEval}
Junyi Li, Xiaoxue Cheng, Xin Zhao, Jian{-}Yun Nie, and Ji{-}Rong Wen. 2023{\natexlab{a}}.
\newblock \href {https://aclanthology.org/2023.emnlp-main.397} {Halueval: {A} large-scale hallucination evaluation benchmark for large language models}.
\newblock In \emph{Proceedings of the 2023 Conference on Empirical Methods in Natural Language Processing, {EMNLP} 2023, Singapore, December 6-10, 2023}, pages 6449--6464. Association for Computational Linguistics.

\bibitem[{Li et~al.(2023{\natexlab{b}})Li, Du, Zhou, Wang, Zhao, and Wen}]{object_hallucination}
Yifan Li, Yifan Du, Kun Zhou, Jinpeng Wang, Wayne~Xin Zhao, and Ji{-}Rong Wen. 2023{\natexlab{b}}.
\newblock \href {https://doi.org/10.48550/arXiv.2305.10355} {Evaluating object hallucination in large vision-language models}.
\newblock \emph{EMNLP}.

\bibitem[{Liang et~al.(2023)Liang, Wu, Song, Wu, Xia, Liu, Ou, Lu, Ji, Mao, Wang, Shou, Gong, and Duan}]{liang2023taskmatrixai}
Yaobo Liang, Chenfei Wu, Ting Song, Wenshan Wu, Yan Xia, Yu~Liu, Yang Ou, Shuai Lu, Lei Ji, Shaoguang Mao, Yun Wang, Linjun Shou, Ming Gong, and Nan Duan. 2023.
\newblock \href {https://doi.org/10.48550/ARXIV.2303.16434} {Taskmatrix.ai: Completing tasks by connecting foundation models with millions of apis}.
\newblock \emph{CoRR}, abs/2303.16434.

\bibitem[{Lightman et~al.(2023)Lightman, Kosaraju, Burda, Edwards, Baker, Lee, Leike, Schulman, Sutskever, and Cobbe}]{lightman2023lets}
Hunter Lightman, Vineet Kosaraju, Yura Burda, Harri Edwards, Bowen Baker, Teddy Lee, Jan Leike, John Schulman, Ilya Sutskever, and Karl Cobbe. 2023.
\newblock \href {http://arxiv.org/abs/2305.20050} {Let's verify step by step}.

\bibitem[{Lin et~al.(2014)Lin, Maire, Belongie, Hays, Perona, Ramanan, Doll{\'a}r, and Zitnick}]{lin2014microsoft}
Tsung-Yi Lin, Michael Maire, Serge Belongie, James Hays, Pietro Perona, Deva Ramanan, Piotr Doll{\'a}r, and C~Lawrence Zitnick. 2014.
\newblock Microsoft coco: Common objects in context.
\newblock In \emph{ECCV}.

\bibitem[{Liu et~al.(2023{\natexlab{a}})Liu, Guan, Li, Chen, Yacoob, Manocha, and Zhou}]{HallusionBench}
Fuxiao Liu, Tianrui Guan, Zongxia Li, Lichang Chen, Yaser Yacoob, Dinesh Manocha, and Tianyi Zhou. 2023{\natexlab{a}}.
\newblock \href {https://doi.org/10.48550/ARXIV.2310.14566} {Hallusionbench: You see what you think? or you think what you see? an image-context reasoning benchmark challenging for gpt-4v(ision), llava-1.5, and other multi-modality models}.
\newblock \emph{CoRR}, abs/2310.14566.

\bibitem[{Liu et~al.(2023{\natexlab{b}})Liu, Lin, Li, Wang, Yacoob, and Wang}]{DBLP:journals/align}
Fuxiao Liu, Kevin Lin, Linjie Li, Jianfeng Wang, Yaser Yacoob, and Lijuan Wang. 2023{\natexlab{b}}.
\newblock \href {https://doi.org/10.48550/ARXIV.2306.14565} {Aligning large multi-modal model with robust instruction tuning}.
\newblock \emph{CoRR}, abs/2306.14565.

\bibitem[{Liu et~al.(2024)Liu, Xue, Chen, Chen, Zhao, Wang, Hou, Li, and Peng}]{liu2024survey}
Hanchao Liu, Wenyuan Xue, Yifei Chen, Dapeng Chen, Xiutian Zhao, Ke~Wang, Liping Hou, Rongjun Li, and Wei Peng. 2024.
\newblock \href {http://arxiv.org/abs/2402.00253} {A survey on hallucination in large vision-language models}.

\bibitem[{Liu et~al.(2023{\natexlab{c}})Liu, Li, Wu, and Lee}]{LLaVA}
Haotian Liu, Chunyuan Li, Qingyang Wu, and Yong~Jae Lee. 2023{\natexlab{c}}.
\newblock \href {https://doi.org/10.48550/ARXIV.2304.08485} {Visual instruction tuning}.
\newblock \emph{CoRR}, abs/2304.08485.

\bibitem[{Liu et~al.(2023{\natexlab{d}})Liu, Zeng, Ren, Li, Zhang, Yang, Li, Yang, Su, Zhu, and Zhang}]{dino23}
Shilong Liu, Zhaoyang Zeng, Tianhe Ren, Feng Li, Hao Zhang, Jie Yang, Chunyuan Li, Jianwei Yang, Hang Su, Jun Zhu, and Lei Zhang. 2023{\natexlab{d}}.
\newblock \href {https://doi.org/10.48550/ARXIV.2303.05499} {Grounding {DINO:} marrying {DINO} with grounded pre-training for open-set object detection}.
\newblock \emph{CoRR}, abs/2303.05499.

\bibitem[{Mishra et~al.(2024)Mishra, Asai, Balachandran, Wang, Neubig, Tsvetkov, and Hajishirzi}]{mishra2024finegrained}
Abhika Mishra, Akari Asai, Vidhisha Balachandran, Yizhong Wang, Graham Neubig, Yulia Tsvetkov, and Hannaneh Hajishirzi. 2024.
\newblock \href {http://arxiv.org/abs/2401.06855} {Fine-grained hallucination detection and editing for language models}.

\bibitem[{OpenAI(2023)}]{gpt4}
OpenAI. 2023.
\newblock Gpt-4 technical report.
\newblock \emph{OpenAI}.

\bibitem[{Peng et~al.(2023)Peng, Galley, He, Cheng, Xie, Hu, Huang, Liden, Yu, Chen, and Gao}]{peng2023check}
Baolin Peng, Michel Galley, Pengcheng He, Hao Cheng, Yujia Xie, Yu~Hu, Qiuyuan Huang, Lars Liden, Zhou Yu, Weizhu Chen, and Jianfeng Gao. 2023.
\newblock \href {https://doi.org/10.48550/ARXIV.2302.12813} {Check your facts and try again: Improving large language models with external knowledge and automated feedback}.
\newblock \emph{CoRR}, abs/2302.12813.

\bibitem[{Qiao et~al.(2023)Qiao, Gui, Chen, and Zhang}]{toolfeed23}
Shuofei Qiao, Honghao Gui, Huajun Chen, and Ningyu Zhang. 2023.
\newblock \href {https://doi.org/10.48550/ARXIV.2305.13068} {Making language models better tool learners with execution feedback}.
\newblock \emph{CoRR}, abs/2305.13068.

\bibitem[{Ramesh et~al.(2022)Ramesh, Dhariwal, Nichol, Chu, and Chen}]{DALLE2}
Aditya Ramesh, Prafulla Dhariwal, Alex Nichol, Casey Chu, and Mark Chen. 2022.
\newblock \href {https://doi.org/10.48550/ARXIV.2204.06125} {Hierarchical text-conditional image generation with {CLIP} latents}.
\newblock \emph{CoRR}, abs/2204.06125.

\bibitem[{Rawte et~al.(2023)Rawte, Sheth, and Das}]{found_hallucination}
Vipula Rawte, Amit~P. Sheth, and Amitava Das. 2023.
\newblock \href {https://doi.org/10.48550/ARXIV.2309.05922} {A survey of hallucination in large foundation models}.
\newblock \emph{CoRR}, abs/2309.05922.

\bibitem[{Saharia et~al.(2022)Saharia, Chan, Saxena, Li, Whang, Denton, Ghasemipour, Lopes, Ayan, Salimans, Ho, Fleet, and Norouzi}]{DrawBench}
Chitwan Saharia, William Chan, Saurabh Saxena, Lala Li, Jay Whang, Emily~L. Denton, Seyed Kamyar~Seyed Ghasemipour, Raphael~Gontijo Lopes, Burcu~Karagol Ayan, Tim Salimans, Jonathan Ho, David~J. Fleet, and Mohammad Norouzi. 2022.
\newblock \href {http://papers.nips.cc/paper\_files/paper/2022/hash/ec795aeadae0b7d230fa35cbaf04c041-Abstract-Conference.html} {Photorealistic text-to-image diffusion models with deep language understanding}.
\newblock In \emph{Advances in Neural Information Processing Systems 35: Annual Conference on Neural Information Processing Systems 2022, NeurIPS 2022, New Orleans, LA, USA, November 28 - December 9, 2022}.

\bibitem[{Schick et~al.(2023)Schick, Dwivedi{-}Yu, Dess{\`{\i}}, Raileanu, Lomeli, Zettlemoyer, Cancedda, and Scialom}]{schick2023toolformer}
Timo Schick, Jane Dwivedi{-}Yu, Roberto Dess{\`{\i}}, Roberta Raileanu, Maria Lomeli, Luke Zettlemoyer, Nicola Cancedda, and Thomas Scialom. 2023.
\newblock Toolformer: Language models can teach themselves to use tools.
\newblock \emph{NeurIPS 2023}.

\bibitem[{Semnani et~al.(2023)Semnani, Yao, Zhang, and Lam}]{semnani2023wikichat}
Sina~J. Semnani, Violet~Z. Yao, Heidi~C. Zhang, and Monica~S. Lam. 2023.
\newblock \href {http://arxiv.org/abs/2305.14292} {Wikichat: Stopping the hallucination of large language model chatbots by few-shot grounding on wikipedia}.

\bibitem[{Shen et~al.(2023)Shen, Song, Tan, Li, Lu, and Zhuang}]{shen2023hugginggpt}
Yongliang Shen, Kaitao Song, Xu~Tan, Dongsheng Li, Weiming Lu, and Yueting Zhuang. 2023.
\newblock Hugginggpt: Solving {AI} tasks with chatgpt and its friends in huggingface.
\newblock \emph{NeurIPS 2023}.

\bibitem[{Shinn et~al.(2023)Shinn, Cassano, Berman, Gopinath, Narasimhan, and Yao}]{shinn2023reflexion}
Noah Shinn, Federico Cassano, Edward Berman, Ashwin Gopinath, Karthik Narasimhan, and Shunyu Yao. 2023.
\newblock \href {http://arxiv.org/abs/2303.11366} {Reflexion: Language agents with verbal reinforcement learning}.

\bibitem[{Singh et~al.(2019)Singh, Natarajan, Shah, Jiang, Chen, Batra, Parikh, and Rohrbach}]{TextVQA}
Amanpreet Singh, Vivek Natarajan, Meet Shah, Yu~Jiang, Xinlei Chen, Dhruv Batra, Devi Parikh, and Marcus Rohrbach. 2019.
\newblock \href {https://doi.org/10.1109/CVPR.2019.00851} {Towards {VQA} models that can read}.
\newblock In \emph{{IEEE} Conference on Computer Vision and Pattern Recognition, {CVPR} 2019, Long Beach, CA, USA, June 16-20, 2019}, pages 8317--8326. Computer Vision Foundation / {IEEE}.

\bibitem[{Tonmoy et~al.(2024)Tonmoy, Zaman, Jain, Rani, Rawte, Chadha, and Das}]{tonmoy2024comprehensive}
S.~M Towhidul~Islam Tonmoy, S~M~Mehedi Zaman, Vinija Jain, Anku Rani, Vipula Rawte, Aman Chadha, and Amitava Das. 2024.
\newblock \href {http://arxiv.org/abs/2401.01313} {A comprehensive survey of hallucination mitigation techniques in large language models}.

\bibitem[{Wan et~al.(2024)Wan, Huang, Cui, Quan, Bi, and Shi}]{KCA_tecent}
Fanqi Wan, Xinting Huang, Leyang Cui, Xiaojun Quan, Wei Bi, and Shuming Shi. 2024.
\newblock \href {https://doi.org/10.48550/ARXIV.2401.10768} {Mitigating hallucinations of large language models via knowledge consistent alignment}.
\newblock \emph{CoRR}, abs/2401.10768.

\bibitem[{Wang et~al.(2020)Wang, Cho, and Lewis}]{wang-etal-2020-asking}
Alex Wang, Kyunghyun Cho, and Mike Lewis. 2020.
\newblock \href {https://doi.org/10.18653/v1/2020.acl-main.450} {Asking and answering questions to evaluate the factual consistency of summaries}.
\newblock In \emph{Proceedings of the 58th Annual Meeting of the Association for Computational Linguistics}, pages 5008--5020, Online. Association for Computational Linguistics.

\bibitem[{Wang et~al.(2023{\natexlab{a}})Wang, Liu, Yue, Tang, Zhang, Cheng, Yao, Gao, Hu, Qi, Wang, Yang, Wang, Xie, Zhang, and Zhang}]{DBLP:journals/xihu/hallu_survey}
Cunxiang Wang, Xiaoze Liu, Yuanhao Yue, Xiangru Tang, Tianhang Zhang, Jiayang Cheng, Yunzhi Yao, Wenyang Gao, Xuming Hu, Zehan Qi, Yidong Wang, Linyi Yang, Jindong Wang, Xing Xie, Zheng Zhang, and Yue Zhang. 2023{\natexlab{a}}.
\newblock \href {https://doi.org/10.48550/ARXIV.2310.07521} {Survey on factuality in large language models: Knowledge, retrieval and domain-specificity}.
\newblock \emph{CoRR}, abs/2310.07521.

\bibitem[{Wang et~al.(2023{\natexlab{b}})Wang, Wang, Xu, Zhang, Gu, Jia, Yan, Zhang, and Sang}]{wang-etl-AMBER}
Junyang Wang, Yuhang Wang, Guohai Xu, Jing Zhang, Yukai Gu, Haitao Jia, Ming Yan, Ji~Zhang, and Jitao Sang. 2023{\natexlab{b}}.
\newblock \href {https://doi.org/10.48550/ARXIV.2311.07397} {An llm-free multi-dimensional benchmark for mllms hallucination evaluation}.
\newblock \emph{CoRR}, abs/2311.07397.

\bibitem[{Wang et~al.(2023{\natexlab{c}})Wang, Zhou, Xu, Shi, Zhao, Xu, Ye, Yan, Zhang, Zhu, Sang, and Tang}]{HaELM}
Junyang Wang, Yiyang Zhou, Guohai Xu, Pengcheng Shi, Chenlin Zhao, Haiyang Xu, Qinghao Ye, Ming Yan, Ji~Zhang, Jihua Zhu, Jitao Sang, and Haoyu Tang. 2023{\natexlab{c}}.
\newblock \href {https://doi.org/10.48550/ARXIV.2308.15126} {Evaluation and analysis of hallucination in large vision-language models}.
\newblock \emph{CoRR}, abs/2308.15126.

\bibitem[{Wang et~al.(2024)Wang, Pan, Ding, and Biemann}]{icd2024}
Xintong Wang, Jingheng Pan, Liang Ding, and Chris Biemann. 2024.
\newblock \href {https://doi.org/10.48550/ARXIV.2403.18715} {Mitigating hallucinations in large vision-language models with instruction contrastive decoding}.
\newblock \emph{CoRR}, abs/2403.18715.

\bibitem[{Wei et~al.(2022)Wei, Wang, Schuurmans, Bosma, Ichter, Xia, Chi, Le, and Zhou}]{wei2023chainofthought}
Jason Wei, Xuezhi Wang, Dale Schuurmans, Maarten Bosma, Brian Ichter, Fei Xia, Ed~H. Chi, Quoc~V. Le, and Denny Zhou. 2022.
\newblock \href {http://papers.nips.cc/paper\_files/paper/2022/hash/9d5609613524ecf4f15af0f7b31abca4-Abstract-Conference.html} {Chain-of-thought prompting elicits reasoning in large language models}.
\newblock In \emph{NeurIPS}.

\bibitem[{Wu et~al.(2024)Wu, Liu, Wang, Zhang, Wu, Wang, and Tan}]{junfei_logical2024}
Junfei Wu, Qiang Liu, Ding Wang, Jinghao Zhang, Shu Wu, Liang Wang, and Tieniu Tan. 2024.
\newblock \href {https://doi.org/10.48550/ARXIV.2402.11622} {Logical closed loop: Uncovering object hallucinations in large vision-language models}.
\newblock \emph{CoRR}, abs/2402.11622.

\bibitem[{Xie et~al.(2023)Xie, Wang, Feng, and Xia}]{xie2023ask}
Qiming Xie, Zengzhi Wang, Yi~Feng, and Rui Xia. 2023.
\newblock \href {https://doi.org/10.48550/ARXIV.2310.02174} {Ask again, then fail: Large language models' vacillations in judgement}.
\newblock \emph{CoRR}, abs/2310.02174.

\bibitem[{Xing et~al.(2024)Xing, Zhao, Wu, An, Chen, Li, Zhang, and Dai}]{EFUF2024}
Shangyu Xing, Fei Zhao, Zhen Wu, Tuo An, Weihao Chen, Chunhui Li, Jianbing Zhang, and Xinyu Dai. 2024.
\newblock \href {https://doi.org/10.48550/ARXIV.2402.09801} {{EFUF:} efficient fine-grained unlearning framework for mitigating hallucinations in multimodal large language models}.
\newblock \emph{CoRR}, abs/2402.09801.

\bibitem[{Yang et~al.(2023)Yang, Pan, Zhao, Chen, Petzold, Wang, and Cheng}]{detect23_survey}
Xianjun Yang, Liangming Pan, Xuandong Zhao, Haifeng Chen, Linda~R. Petzold, William~Yang Wang, and Wei Cheng. 2023.
\newblock \href {https://doi.org/10.48550/ARXIV.2310.15654} {A survey on detection of llms-generated content}.
\newblock \emph{CoRR}, abs/2310.15654.

\bibitem[{Ye et~al.(2023)Ye, Xu, Xu, Ye, Yan, Zhou, Wang, Hu, Shi, Shi, Li, Xu, Chen, Tian, Qi, Zhang, and Huang}]{mPLUG-Owl}
Qinghao Ye, Haiyang Xu, Guohai Xu, Jiabo Ye, Ming Yan, Yiyang Zhou, Junyang Wang, Anwen Hu, Pengcheng Shi, Yaya Shi, Chenliang Li, Yuanhong Xu, Hehong Chen, Junfeng Tian, Qian Qi, Ji~Zhang, and Fei Huang. 2023.
\newblock \href {https://doi.org/10.48550/ARXIV.2304.14178} {mplug-owl: Modularization empowers large language models with multimodality}.
\newblock \emph{CoRR}, abs/2304.14178.

\bibitem[{Yin et~al.(2023)Yin, Fu, Zhao, Xu, Wang, Sui, Shen, Li, Sun, and Chen}]{Woodpecker}
Shukang Yin, Chaoyou Fu, Sirui Zhao, Tong Xu, Hao Wang, Dianbo Sui, Yunhang Shen, Ke~Li, Xing Sun, and Enhong Chen. 2023.
\newblock \href {https://doi.org/10.48550/ARXIV.2310.16045} {Woodpecker: Hallucination correction for multimodal large language models}.
\newblock \emph{CoRR}, abs/2310.16045.

\bibitem[{Zhai et~al.(2023)Zhai, Yang, Zhao, Xu, Shen, Zhao, Keutzer, Li, Yan, and Fan}]{zhai2023halle}
Bohan Zhai, Shijia Yang, Xiangchen Zhao, Chenfeng Xu, Sheng Shen, Dongdi Zhao, Kurt Keutzer, Manling Li, Tan Yan, and Xiangjun Fan. 2023.
\newblock \href {https://doi.org/10.48550/ARXIV.2310.01779} {Halle-switch: Rethinking and controlling object existence hallucinations in large vision language models for detailed caption}.
\newblock \emph{CoRR}, abs/2310.01779.

\bibitem[{Zhang et~al.(2023{\natexlab{a}})Zhang, Cui, Bi, and Shi}]{induce2024}
Yue Zhang, Leyang Cui, Wei Bi, and Shuming Shi. 2023{\natexlab{a}}.
\newblock \href {https://doi.org/10.48550/ARXIV.2312.15710} {Alleviating hallucinations of large language models through induced hallucinations}.
\newblock \emph{CoRR}, abs/2312.15710.

\bibitem[{Zhang et~al.(2023{\natexlab{b}})Zhang, Li, Cui, Cai, Liu, Fu, Huang, Zhao, Zhang, Chen, Wang, Luu, Bi, Shi, and Shi}]{suvery-Siren}
Yue Zhang, Yafu Li, Leyang Cui, Deng Cai, Lemao Liu, Tingchen Fu, Xinting Huang, Enbo Zhao, Yu~Zhang, Yulong Chen, Longyue Wang, Anh~Tuan Luu, Wei Bi, Freda Shi, and Shuming Shi. 2023{\natexlab{b}}.
\newblock \href {https://doi.org/10.48550/arXiv.2309.01219} {Siren's song in the {AI} ocean: {A} survey on hallucination in large language models}.
\newblock \emph{CoRR}, abs/2309.01219.

\bibitem[{Zhao et~al.(2024)Zhao, Deng, Zhang, and Gu}]{MARINE}
Linxi Zhao, Yihe Deng, Weitong Zhang, and Quanquan Gu. 2024.
\newblock \href {https://doi.org/10.48550/ARXIV.2402.08680} {Mitigating object hallucination in large vision-language models via classifier-free guidance}.
\newblock \emph{CoRR}, abs/2402.08680.

\bibitem[{Zhou et~al.(2023)Zhou, Cui, Yoon, Zhang, Deng, Finn, Bansal, and Yao}]{LURE}
Yiyang Zhou, Chenhang Cui, Jaehong Yoon, Linjun Zhang, Zhun Deng, Chelsea Finn, Mohit Bansal, and Huaxiu Yao. 2023.
\newblock \href {https://doi.org/10.48550/ARXIV.2310.00754} {Analyzing and mitigating object hallucination in large vision-language models}.
\newblock \emph{CoRR}, abs/2310.00754.

\bibitem[{Zhu et~al.(2023)Zhu, Chen, Shen, Li, and Elhoseiny}]{MiniGPT4}
Deyao Zhu, Jun Chen, Xiaoqian Shen, Xiang Li, and Mohamed Elhoseiny. 2023.
\newblock \href {https://doi.org/10.48550/ARXIV.2304.10592} {Minigpt-4: Enhancing vision-language understanding with advanced large language models}.
\newblock \emph{CoRR}, abs/2304.10592.

\end{thebibliography}

\appendix

\section{Prompt Templates}
\label{sec:app_prompt}
Within this section, we outline the prompt templates designed to guide the foundational MLLM  for the autonomous  query 
formulation (illustrated in Table~\ref{tab:query_object}-\ref{tab:query_fact}) and verification of any hallucinated content (shown in Table~\ref{tab:prompt_i2t}-\ref{tab:prompt_t2i}).


\begin{table*}[!ht]\centering
\begin{minipage}{0.85\textwidth}
\centering
\begin{tcolorbox} 
    \centering
   
      \small
    \begin{tabular}{p{0.95\textwidth}}
   \textcolor[rgb]{0.8,0.3,0}{ {\bf SYSTEM:} } \\ 
   You are a brilliant object extractor. \\ \\
  \textcolor[rgb]{0.8,0.3,0}{ {\bf USER:} } \\ 
    Given a list of claim, extract the objects from each claim for me. \\
      Extract the common objects and summarize them as general categories without repetition, merge essentially similar objects. \\
      Avoid extracting hypernyms, keep hyponyms! \\
      Avoid extracting abstract or non-specific objects.  \\
      Extract object in the singular form. \\
      Output all the extracted types of items separate each object type with a period. \\
      If there is nothing to output, then output a single "none". \\
      YOU MUST TO DISREGARD OBJECT WORDS THAT ARE NOT NATURAL OBJECTS, SUCH AS SCENES, AREA, SKY, GROUND, WORDS, ATMOSPHERES, COUNTRIES, NAMES, AND PLACES.IF THERE ARE NO NATURAL objects IN THE SENTENCE, RETURN 'none'. \\
      YOU MUST RETURN THE RESULTS IN A DICTIONARY ACCORDING TO THE GIVEN ORDER OF THE LIST OF CLAIMS. \\
      You MUST only respond in the format as described below. DO NOT RESPOND WITH ANYTHING ELSE. \\
      
      response format: \{\{"claim1":"object1.object2.object3","claim2":"none","claim3":"object1.object2", ...\}\} \\ \\

      Here are three examples: \\
      claim list: \\
      claim1: The image depicts a man laying on the ground. \\
      claim2: The man is next to a motorcycle. \\
      claim3: The sun is shining upon the ground. \\
      claim4: The light is very bright. \\
      output: \\
      \{\{"claim1":"man","claim2":"man.motorcycle","claim3":"none", "claim4":"none"\}\}
\\ \\
      claim list: \\ 
      claim1: The image shows a device. \\
      claim2: The device has the words  \"Samsung\". \\
      claim3: Samsung is a Korean company. \\
      output: \\
      \{\{"claim1":"device","claim2":"device", "claim3":"none"\}\} \\ \\

      claim list: \\
      claim1: A man wears a green shirt. \\
      claim2: The man's face is beaming with a smile. \\
      claim3: The image shows the man in high spirits. \\
      output: \\    \{\{"claim1":"man.shirt","claim2":"man","claim3":"man"\}\} \\ \\

      Now complete your output with following the above rules. \\
      claim list: \\
      \{claims\} \\
      output:

    \end{tabular}
\end{tcolorbox}
\caption{Prompt template of query formulation (object-level) for image-to-text generation.}
\label{tab:query_object}
\end{minipage}
\end{table*}

\begin{table*}[!ht]\centering
\begin{minipage}{0.85\textwidth}
\centering
\begin{tcolorbox} 
    \centering
   
      \small
    \begin{tabular}{p{0.95\textwidth}}
   \textcolor[rgb]{0.8,0.3,0}{ {\bf SYSTEM:} } \\ 
        You are a brilliant question generator. \\ \\
  \textcolor[rgb]{0.8,0.3,0}{ {\bf USER:} } \\ 
    Given a list of claim and some objects(each object is connected by a period), you're required to generate questions about attributes of the given objects.  \\
      The generated questions may involve basic attributes such as colors, actions and position mentioned in the claim.  \\
      Do not ask questions involving object counts or the existence of object.
      Do not ask questions involving scene text. \\
      When asking questions about attributes, try to ask simple questions that only involve one object. \\
      Ask questions that can be easily decided visually. Do not ask questions that require complex reasoning. \\
      Do not ask semantically similar questions. Do not ask questions only about scenes or places. \\
      Do not ask questions about uncertain or conjecture parts of the claim, for example, the parts described with "maybe" or "likely", etc. \\
      It is no need to cover all the specified objects.
      If there is no question to ask, simply output 'none'. \\
      YOU MUST RETURN THE RESULTS IN A DICTIONARY ACCORDING TO THE GIVEN ORDER OF THE LIST OF CLAIMS. \\
      You MUST only respond in the format as described below. 
      DO NOT RESPOND WITH ANYTHING ELSE. \\
      \\
      response format: \{\{"claim1":["question1", "question2"],"claim2":["none"],"claim3":["question1", "question2"], ...\}\} \\ \\

     Here are three examples: \\ 
     objects:\\
      dog.cat \\
      claim list: \\
      claim1: There is one black dog on the left in the image. \\
      claim2: There are two white cats on the right in the image. \\
      output: \\
      \{\{"claim1":["What color is the dog?", "Is there a dog on the left in the image?"],"claim2":["What color are the cat?", "Are there two cats on the right in the image?"]\}\} \\ \\

   objects: \\
      man.baseball cap.wall \\
      claim list: \\
      claim1: The man is wearing a baseball cap. \\
      claim2: The man appears to be smoking. \\
      claim3: 'hello world' is written on the white wall. \\
      output: \\
      \{\{"claim1":["What is the man wearing?"], "claim2":["Does the man appear to be smoking?"], "claim3":[What color is the wall?]\}\} \\ \\

      objects: \\
      kitchen.man.apron \\
      claim list: \\
      claim1: The image depicts a kitchen. \\
      claim2: There is a man in a white apron.  \\
      claim3: The man is standing in the middle of the kitchen. \\
      claim4: The overall atmosphere is very pleasant. \\
      output: \\
      {{"claim1":["none"], "claim2":["What does the man wear?", "What color is the apron?"], "claim3":["Is the man standing in the middle of the kitchen?"], "claim4": ["none"]}}  \\ \\

      Now complete the following with following the above rules. DO NOT RESPOND WITH ANYTHING ELSE. \\
      objects:  \\
      \{objects\} \\
      claim list: \\
      \{claims\} \\
      output:

    \end{tabular}
\end{tcolorbox}
\caption{Prompt template of query formulation (attribute-level) for image-to-text generation.}
\label{tab:query_attribute}
\end{minipage}
\end{table*}

\begin{table*}[!ht]\centering
\begin{minipage}{0.85\textwidth}
\centering
\begin{tcolorbox} 
    \centering
   
      \small
    \begin{tabular}{p{0.95\textwidth}}
   \textcolor[rgb]{0.8,0.3,0}{ {\bf SYSTEM:} } \\ 
        You are a brilliant question generator. \\ \\
  \textcolor[rgb]{0.8,0.3,0}{ {\bf USER:} } \\ 
Given a list of claim, you're required to generate questions about scene text to assist users in verifying the accuracy of the claim.  \\
      If the information mentioned in this claim pertains to scene text, you'll need to generate question about the scene text. \\
      If the claim is unrelated to the scene text information in the image, such as: objects, colors, actions, position etc, simply return 'none'. \\
      YOU MUST RETURN THE RESULTS IN A DICTIONARY ACCORDING TO THE GIVEN ORDER OF THE LIST OF CLAIMS. \\
      You MUST only respond in the format as described below. DO NOT RESPOND WITH ANYTHING ELSE.\\ 
      response format: \{\{"claim1":["question1", "question2"],"claim2":["none"],"claim3":["question1", "question2"], ...\}\} \\ \\

      Here are three examples: \\
      claim list: \\
      claim1: There is a black device in the image. \\
      claim2: The device is a brand of smartphones produced by Samsung Electronics. \\
      output: \{\{"claim1":["none"],"claim2":["What is the brand of the device in the image?"]\}\} \\ \\
      
      claim list: \\
      claim1: A stop sign is on the left. \\
      claim2: The stop sign says stop eating animals. \\
      output: \{\{"claim1":["none"],"claim2":["What does the stop sign say in the image?"]\}\} \\ \\

      claim list: \\
      claim1: The words 'Hello World' are written on the car. \\
      claim2: A man is standing beside the car. \\
      output: \{\{"claim1":["What are written on the car?"],"claim2":["none"]\}\} \\ \\

      Now complete the following with following the above rules. DO NOT RESPOND WITH ANYTHING ELSE. \\
      claim list:  \\
      \{claims\} \\
      output:

    \end{tabular}
\end{tcolorbox}
\caption{Prompt template of query formulation (scene-text-level) for image-to-text generation.}
\label{tab:query_scene}
\end{minipage}
\end{table*}

\begin{table*}[!ht]\centering
\begin{minipage}{0.85\textwidth}
\centering
\begin{tcolorbox} 
    \centering
   
      \small
    \begin{tabular}{p{0.95\textwidth}}
   \textcolor[rgb]{0.8,0.3,0}{ {\bf SYSTEM:} } \\ 
        You are a brilliant question generator. \\ \\
  \textcolor[rgb]{0.8,0.3,0}{ {\bf USER:} } \\ 
Given a list of claim, you're required to generate questions about related to factual visual information. \\
      For a claim based on factual knowledge, Your primary task is to generate a Python list of two effective and skeptical search engine questions. \\
      These questions should assist users in critically evaluating the factuality of a provided claim using search engines. \\
      If a claim is not based on factual knowledge, simply return 'none'. \\
      YOU MUST RETURN THE RESULTS IN A DICTIONARY ACCORDING TO THE GIVEN ORDER OF THE LIST OF CLAIMS. \\
      You MUST only respond in the format as described below. DO NOT RESPOND WITH ANYTHING ELSE. \\
      response format: \{\{"claim1":["question1", "question2"],"claim2":["none"],"claim3":["question1", "question2"], ...\}\} \\ \\
      
 Here are three examples: \\
        claim list: \\
      claim1: The image shows a black phone. \\
      claim2: This black phone is manufactured by Huawei. \\
      claim3: Huawei is a company located in Shenzhen, China. \\
      output:  \\
      \{\{"claim1":["none"],"claim2":["none"],"claim3":["Where is Huawei headquartered?", "Huawei company"]\}\} \\ \\

      claim list: \\
      claim1: The image shows an app of twitter. \\
      claim2: The CEO of twitter is Bill Gates. \\
      output: \{\{"claim1":["none"],"claim2":["Who is the CEO of twitter?", "CEO Twitter"]\}\} \\ \\

      claim list: \\
      claim1: The man is playing baseball. \\
      claim2: The man is wearing a colorful shirt. \\
      output: \{\{"claim1":["none"],"claim2":["none"]\}\} \\ \\

       Now complete the following with following the above rules. DO NOT RESPOND WITH ANYTHING ELSE. \\
      claim list: \\
     \{claims\} \\
      output:

    \end{tabular}
\end{tcolorbox}
\caption{Prompt template of query formulation (fact-level) for image-to-text generation.}
\label{tab:query_fact}
\end{minipage}
\end{table*}

\begin{table*}[!ht]\centering
\begin{minipage}{0.85\textwidth}
\centering
\begin{tcolorbox} 
    \centering
   
      \small
    \begin{tabular}{p{0.95\textwidth}}
   \textcolor[rgb]{0.8,0.3,0}{ {\bf SYSTEM:} } \\ 
   You are a brilliant hallucination judger. \\ \\
  \textcolor[rgb]{0.8,0.3,0}{ {\bf USER:} } \\ 
  Given a list of claims from Multimodal Large Language Models and an image, you are required to judge whether each claim in the list by the Multimodal Large Language Model model conflicts with the image, following these rules: 

    1. You must carefully judge from four aspects, including the object, attributes, scene text and fact. Here are specific descriptions of the four aspects for you to review: \\
        "Object" specifically refers to whether the objects in the image exist and if the quantity of objects conflicts with the object information in the claims; \\
        "Attributes" specifically refer to whether the color, position, action of objects in the image conflict with the attribute information in the claims; \\
        "Scene Text" specifically refers to whether the textual information in the scene of the image conflicts with the required textual information in the claims. \\
        "Fact" specifically refers to relevant factual knowledge obtained by querying a search engine. You can verify the factual accuracy of the claims based on the provided external knowledge. \\
        
    2. You'll also receive detection results from the expert model. 
        The object detection expert model will provide detected entity names along with their bounding box information in the image. When deriving position relationships between entity instances, try to also use the bounding boxes information, which are represented as [x1, y1, x2, y2] with floating numbers ranging from 0 to 1. These values correspond to the top left x1, top left y1, bottom right x2, and bottom right y2. 
        The scene text expert model will provide detected specific text along with their bounding box information in the image. As long as there is a conflict between a single letter in the scene text and the text information required in the claim, it's considered a hallucination.
        
    3. You must carefully judge whether the visual information in the image conflicts with each claim. If there is a conflict, the result for that statement is labeled as 'hallucination'; otherwise, it is labeled as 'non-hallucination'."
        
    4. Finally, YOU MUST RETURN THE JUDGMENT RESULTS IN A DICTIONARY ACCORDING TO THE GIVEN ORDER OF THE LIST OF CLAIMS. You MUST only respond in the format as described below. DO NOT RESPOND WITH ANYTHING ELSE.
        response format: [{"claim1":"hallucination", "reason":"The reason for your judgment."},{"claim2":"non-hallucination", "reason":"The reason for your judgment."},{"claim3":"hallucination", "reason":"The reason for your judgment."}, ...]

\\
\\
\textcolor[rgb]{0,0.7,0}{ \lbrack{}Begin of Example \rbrack{} (Image Entered)}  \\
 Here is the object detection expert model's result: \\
    people [0.345, 0.424, 0.408, 0.509];
    people [0.197, 0.44, 0.28, 0.514] \\
    people [0.517, 0.315, 0.561, 0.401];
    people [0.441, 0.356, 0.47, 0.405] \\
    chair [0.398, 0.595, 0.637, 0.901];
    chair [0.621, 0.592, 0.789, 0.889] \\
    umbrella [0.501, 0.334, 0.968, 0.88] \\
\\
Here is the attribute detection expert model's result:
    none information \\ \\
    
    Here is the scene text recognition expert model's result: 
    none information 
\\  \\
    Here is the external knowledge:
    none information \\ \\

    Here is the claim list: \\
    claim1: The picture shows five people swimming. \\
    claim2: On the beach, there is a chair, a umbrella, and a surfboard. \\
    claim3: The green umbrella is on the right side of the chair. \\ \\

    Output:
    [
        {"claim1":"hallucination","reason":"The object detection expert model identified four people, not five people. Based on the image information, they might be swimming. Therefore, there's a hallucination."},
        {"claim2":"hallucination","reason":"According to the results of the object detection expert model and my judgment, there are two chairs and an umbrella in the picture, but there is no surfboard. Therefore, there's a hallucination."},
        {"claim3":"non-hallucination","reason":"Based on the positional information of the bounding boxes and my judgment, the umbrella is to the right of the chairs. The umbrella is green. Therefore, there's no hallucination."}
    ]
\\
...... \\
\textcolor[rgb]{0,0.7,0}{ \lbrack{}End of Example \rbrack{} }  \\
\\

   \textcolor[rgb]{0.8,0.3,0}{ {\bf <Input>:}  } \\ 
   \textcolor[rgb]{0.8,0.3,0}{ {\bf <Output>:}  } \\ 
    \end{tabular}
\end{tcolorbox}
\caption{Prompt template of hallucination verification for image-to-text generation.}
\label{tab:prompt_i2t}
\end{minipage}
\end{table*}

\begin{table*}[!ht]\centering
\begin{minipage}{0.9\textwidth}
\centering
\begin{tcolorbox} 
    \centering
   
      \small
    \begin{tabular}{p{0.95\textwidth}}
   \textcolor[rgb]{0.8,0.3,0}{ {\bf SYSTEM:} } \\ 
   You are a brilliant hallucination judger. \\ \\
  \textcolor[rgb]{0.8,0.3,0}{ {\bf USER:} } \\ 
  Given a list of claims from human prompts, an image generated by the text-to-image model, you are required to judge whether the image conflicts with human-provided prompts, following these rules:

    1. You must carefully judge from four aspects, including the object, attributes, scene text and fact. Here are specific descriptions of the four aspects for you to review: \\
        "Object" specifically refers to whether the objects in the image exist and if the quantity of objects conflicts with the object information in the claims; \\
        "Attributes" specifically refer to whether the color, position, action of objects in the image conflict with the attribute information in the claims; \\
        "Scene Text" specifically refers to whether the textual information in the scene of the image conflicts with the required textual information in the claims. \\
        "Fact" specifically refers to relevant factual knowledge obtained by querying a search engine. You can verify the factual accuracy of the claims based on the provided external knowledge. \\
        
    2. You'll also receive detection results from the expert model. 
        The object detection expert model will provide detected entity names along with their bounding box information in the image. When deriving position relationships between entity instances, try to also use the bounding boxes information, which are represented as [x1, y1, x2, y2] with floating numbers ranging from 0 to 1. These values correspond to the top left x1, top left y1, bottom right x2, and bottom right y2. 
        The scene text expert model will provide detected specific text along with their bounding box information in the image. As long as there is a conflict between a single letter in the scene text and the text information required in the claim, it's considered a hallucination.
        
    3. You must carefully judge whether the visual information in the image conflicts with each claim. If there is a conflict, the result for that statement is labeled as 'hallucination'; otherwise, it is labeled as 'non-hallucination'."
        
    4. Finally, YOU MUST RETURN THE JUDGMENT RESULTS IN A DICTIONARY ACCORDING TO THE GIVEN ORDER OF THE LIST OF CLAIMS. You MUST only respond in the format as described below. DO NOT RESPOND WITH ANYTHING ELSE.
        response format: [{"claim1":"hallucination", "reason":"The reason for your judgment."},{"claim2":"non-hallucination", "reason":"The reason for your judgment."},{"claim3":"hallucination", "reason":"The reason for your judgment."}, ...]

\\
\\
\textcolor[rgb]{0,0.7,0}{ \lbrack{}Begin of Example \rbrack{} (Image Entered)}  \\

  Here is the object detection expert model's result: \\
    basketball [0.741, 0.179, 0.848, 0.285] \\
    boy [0.773, 0.299, 0.98, 0.828] \\ 
    car [0.001, 0.304, 0.992, 0.854] \\ \\

Here is the attribute detection expert model's result:
    none information \\ \\
    
    Here is the scene text recognition expert model's result: \\
    worlld [0.405, 0.504, 0.726, 0.7] \\ \\

    Here is the external knowledge:
    none information \\ \\

    Here is the claim list: \\
    claim1: The side of the car reads 'Hello World' \\
    claim2: A boy is playing a yellow basketball beside a plant. \\ \\

    Output: [{"claim1":"hallucination", "reason":"The object detection model has identified a car in the image. However, based on the detection results of the scene text expert model and my judgment, the text in the image is 'hello worlld' not 'hello world'. Therefore, there's a hallucination."},{"claim2":"hallucination", "reason":"The object detection model has identified a boy and a basketball in the image. And the boy is visible in the image playing with a yellow basketball. But according to the detection results of the object detection expert model and my judgment, there's no plant. Therefore, there's a hallucination."}]
\\
...... \\
\textcolor[rgb]{0,0.7,0}{ \lbrack{}End of Example \rbrack{} }  \\
\\

   \textcolor[rgb]{0.8,0.3,0}{ {\bf <Input>:}  } \\ 
   \textcolor[rgb]{0.8,0.3,0}{ {\bf <Output>:}  } \\ 
    \end{tabular}
\end{tcolorbox}
\caption{Prompt template of hallucination verification for text-to-image generation.}
    \label{tab:prompt_t2i}
\end{minipage}
\end{table*}

\end{document}